\documentclass[3p,twocolumn]{elsarticle}
\usepackage{geometry}
\usepackage{amsmath,amsfonts}
\usepackage{algorithmic}
\usepackage{algorithm}
\usepackage{array}
\usepackage[caption=false,font=normalsize,labelfont=sf,textfont=sf]{subfig}
\usepackage{textcomp}
\usepackage{stfloats}
\usepackage{url}
\usepackage{verbatim}
\usepackage{graphicx}
\usepackage{times}
\usepackage{epsfig}
\usepackage{wrapfig}
\usepackage{amsfonts}
\usepackage{mathtools}
\usepackage{multirow}
\usepackage{gensymb}

\newcommand{\squeezelittle}{\vspace{-1mm}}
\usepackage{xcolor}
\newcommand{\Tau}{\mathrm{T}}
\usepackage{pifont}
\usepackage{color, colortbl}
\definecolor{brightgreen}{rgb}{0.4, 1.0, 0.0}
\definecolor{LightCyan}{rgb}{0.88,1,1}
\definecolor{Gray}{gray}{0.9}
\definecolor{darkgreen}{rgb}{0, 0.56, 0.3176}
\definecolor{darkred}{rgb}{0.933, 0.133, 0.047}
\definecolor{darkorange}{rgb}{1., 0.576, 0.}
\newcommand*\colourcheck[1]{%
  \expandafter\newcommand\csname #1check\endcsname{\textcolor{#1}{\ding{52}}}%
}
\newcommand*\colourx[1]{%
  \expandafter\newcommand\csname #1x\endcsname{\textcolor{#1}{\ding{55}}}%
}
\colourcheck{blue}
\colourcheck{green}
\colourcheck{red}
\colourcheck{brightgreen}
\colourx{blue}
\colourx{green}
\colourx{red}
\usepackage[utf8]{inputenc} % allow utf-8 input
\usepackage[T1]{fontenc}    % use 8-bit T1 fonts
\usepackage{hyperref}       % hyperlinks
\usepackage{url}            % simple URL typesetting
\usepackage{booktabs}       % professional-quality tables
\usepackage{amsfonts}       % blackboard math symbols
\usepackage{nicefrac}       % compact symbols for 1/2, etc.
\usepackage{microtype}      % microtypography
\usepackage{xcolor}         % colors

\usepackage{amssymb}
\usepackage{amsmath}
\journal{Knowledge-Based Systems}

\begin{document}

\begin{frontmatter}



\title{Semantic Environment Atlas for Object-Goal Navigation}

% \affiliation[label1]{organization={},

% \affiliation[label2]{organization={},

% \author[label1]{Nuri Kim, Jeongho Park, Mineui Hong, Songhwai Oh} %% Author name

\author{{Nuri Kim}}
\ead{nuri.kim@rllab.snu.ac.kr}

\author{{Jeongho Park}}
\ead{jeongho.park@rllab.snu.ac.kr}

\author{{Mineui Hong}}
\ead{mineui.hong@rllab.snu.ac.kr}

\author{{Songhwai Oh}\corref{cor}}
\cortext[cor]{Corresponding author.}
\ead{songhwai@snu.ac.kr}

\affiliation[]{organization={Department of Electrical and Computer Engineering and ASRI, Seoul National University},%Department and Organization
            addressline={1 Gwanak-ro, Gwanak-gu}, 
            city={Seoul},
            postcode={08826},
            country={Republic of Korea}}
%% Abstract
\begin{abstract}
In this paper, we introduce the Semantic Environment Atlas (SEA), a novel mapping approach designed to enhance visual navigation capabilities of embodied agents. The SEA utilizes semantic graph maps that intricately delineate the relationships between places and objects, thereby enriching the navigational context. These maps are constructed from image observations and capture visual landmarks as sparsely encoded nodes within the environment. The SEA integrates multiple semantic maps from various environments, retaining a memory of place-object relationships, which proves invaluable for tasks such as visual localization and navigation. We developed navigation frameworks that effectively leverage the SEA, and we evaluated these frameworks through visual localization and object-goal navigation tasks. Our SEA-based localization framework significantly outperforms existing methods, accurately identifying locations from single query images. Experimental results in Habitat~\cite{habitat19iccv} scenarios show that our method not only achieves a success rate of  39.0\%—an improvement of 12.4\% over the current state-of-the-art—but also maintains robustness under noisy odometry and actuation conditions, all while keeping computational costs low.
% In this paper, we propose the semantic environment atlas (SEA), a novel mapping approach designed for the visual navigation of embodied agents. The key idea of SEA is the development of a rich semantic graph map that intricately delineates the relationships between places and objects, thereby enriching the navigational context for agents. 
% % 
% Firstly, semantic graph maps are collected to build the SEA.
% % 
% The semantic graph, built from image observations, includes sparsely encoded place and object nodes. The graph captures the visual landmarks of the environment. 
% % 
% Then, SEA is constructed from multiple semantic maps from different environments. It has a place-object relationship as a memory.
% % 
% By incorporating semantic knowledge, the SEA becomes a valuable guide for embodied agents performing visual localization and navigation tasks. 
% % 
% We propose navigation frameworks that effectively utilize the SEA. 
% % 
% We tested our frameworks on visual localization and a downstream navigation task, e.g., object-goal navigation.
% % 
% The SEA-based localization framework accurately identifies location from a single query image, outperforming other methods. 
% % 
% Experimental results show that our method is robust under noisy odometry and actuation conditions, while also keeping computational costs low. In Habitat~\cite{habitat19iccv} object-goal navigation scenarios, our navigation framework achieves a 39.0\% success rate, representing a 12.4\% improvement over the current state-of-the-art.
\end{abstract}


%% Keywords

\end{frontmatter}



\section{Introduction}
\label{sec:intro}
Embodied AI technologies, which are becoming increasingly ubiquitous in modern life, are proving integral to various applications, including delivery robots, household chore robots, and self-driving cars.
The pivotal success factor in this field has been the development of intelligent agents that use RGB sensors to interpret semantic knowledge, particularly through learning-based methods such as reinforcement learning (RL)~\cite{mirowski2018learning, chen2019behavioral, yang2019embodied, Exp4nav, kumar2018visual, SMT, SPTM, gupta2017cognitive, parisotto2017neural, avraham2019empnet, lv2020improving, objectGraphNavi, qiu2020target, ANS, SemExp, Online, TSGM, VGM, THDA, PONI, RedRabbit, bayesian}.
% However, these learning-based methods, while powerful, introduce a substantial challenge: high computational costs. 
These methods, while powerful, introduce a significant challenge: high computational costs.

Addressing this challenge, this paper introduces the Semantic Environment Atlas (SEA), a novel map type. The SEA is specifically designed to tackle visual localization and navigation tasks in a computationally efficient manner. The SEA sets itself apart with three distinctive characteristics that collectively enable successful visual navigation.
% In this paper, we propose a semantic environment atlas (SEA), a novel map type for tackling visual localization and navigation tasks, enabling navigation task with low computational costs. 
% The SEA has three distinct characteristics that facilitate successful visual navigation. 

The first distinctive characteristic of the SEA is its \textit{robust} navigation performance against sensor noise. Sensor noise is a common problem in navigation tasks, which tends to accumulate during sequential decision-making processes. Traditional approaches have tried to mitigate this issue through loop closure, but such solutions are challenging for deep learning-based methods that lack state-space-based noise filtering. Consequently, current navigation methods~\cite{SemExp, PONI, ANS, FBE} often assume a noiseless pose sensor—an unrealistic premise in real-world scenarios. In contrast, our method leverages semantic knowledge, enabling it to navigate robustly even with noisy sensors.

% Firstly, SEA is robust to sensor noises. The sensor noise accumulates in sequential decision-making processes. To address this issue, the conventional methods [ref.] commonly utilized loop closure to reduce the error. Since deep learning-based methods do not perform state-space-based noise filtering, it has difficulty introducing the deep-learning based method. Therefore, deep learning-based object-goal navigation methods~\cite{SemExp, PONI, ANS, FBE} assumed there is noiseless pose sensor, which is an unrealistic assumption. Using semantic knowledge, our method can successfully localize the robot location, and robustly  navigate with noisy sensors.

The second distinctive property of the SEA is its ability to \textit{localize} the current position using semantic knowledge. This capability addresses a key challenge: predicting an object’s position with a partially observed map. While recent work~\cite{Online, OTZ,NRNS} has integrated graph-based priors into the metric map to counter this issue, our method takes a step further by incorporating additional semantic knowledge, such as the relationships between objects and places, thereby bolstering localization performance.

% Secondly, SEA can \textit{localize} the current position by utilizing semantic knowledge. Additionally, it is challenging to predict the position of an object with a partially observed map. To address this issue, recent papers~\cite{Online, OTZ,NRNS} integrate graph-based priors into the metric map. Our method integrates more semantic knowledge like relationship between objects and places and improved localization performance.

The third and final property of the SEA is its \textit{adaptability}. Unlike recent methods~\cite{ANS,SemExp} which do not update upon environmental changes, the SEA is designed to self-update based on these changes. This adaptive quality permits navigation agents to adjust their destinations and explore alternative target locations if the initial object search is unsuccessful.
% Lastly, SEA shows a remarkable resilience to new environment by adaptively updating its semantic knowledge. Traditional map types, such as graph maps and metric maps, which store keyframes as nodes and objects in a grid map, respectively, do not update the map when the graph is constructed due to environmental changes. Instead, SEA updates by itself based on the changes in the environment. This allows navigation agents to adjust their destinations based on the updated graph map if they fail to locate the intended object, thus enabling the exploration of alternative target locations.
% For instance, in the task of identifying a cup, a navigation agent will initially search the kitchen, where cups are typically found. However, if the agent reaches the kitchen and cannot locate the cup, the correlation between the kitchen and cup in SEA will decrease, and the bathroom having the highest probability of containing the cup will be the new target location.

The SEA is constructed using semantic graph maps, which incorporate both place-object and place-place relationships. An agent uses the place-object relationship to pinpoint the target location where an object is most likely to be found. To reach this target, the agent leverages place relationships to determine the optimal semantic path. For local navigation, the agent identifies subgoal candidates based on current object observations and chooses the subgoal with the highest reachability to the target. Our method's reliance on semantic path planning eliminates the need for a global pose sensor, thus enhancing robustness against noisy odometry sensors. Additionally, we implement relation updates in new environments since the semantic structure can vary significantly from one environment to another. These updates allow the place graph to adapt, ensuring the most effective semantic path is selected in new settings.
% The SEA is generated using semantic graph maps and consists of the place-object relationship and place relationship. Using the place-object relationship, an agent can find the target place where the object is most likely to be located. {\color{black}To navigate to the target place}, place relationship is utilized to estimate the best semantic path. For local navigation, the agent extracts subgoal candidates based on the current object observation and chooses the best subgoal with the highest reachability to the target place. Since we use semantic path planning, {\color{black}the} global pose sensor becomes unnecessary, which makes our method robust to noisy odometry sensors. Additionally, we employ relation updates on a new environment, since the semantic structure of the environment can differ from the seen environments. Using the relation update, the place graph can be adapted to the new environment to find the best semantic path for the new environment. 
% We have evaluated the proposed method to object goal navigation using MP3D \cite{Matterport3D}. 

The proposed method is evaluated using MP3D~\cite{Matterport3D} for object goal navigation. The experimental study demonstrated that our navigation framework, by leveraging the SEA, achieved a success rate of 39.0\%. This result marks a substantial 12.4\% improvement over the current state-of-the-art.
% , thereby highlighting the SEA's potential to improve the field of semantic knowledge-based embodied agent navigation.
% The SEA is generated using semantic graph maps and consists of the place-object relationship and place relationship. Using the place-object relationship, an agent can find the target place where the object is most likely to be located. To navigate the target place, place relationship is utilized to estimate the best semantic path. For local navigation, the agent extracts subgoal candidates based on the current object observation and chooses the best subgoal with the highest reachability to the target place. Since we use semantic path planning, global pose sensor becomes unnecessary, which makes our method robust to noisy odometry sensors. Additionally, we employ relation updates on test episodes since the test environment may differ from the training environment. Using the relation update, the posterior place graph is generated to find the best semantic path. We have evaluated the proposed method to object goal navigation using MP3D \cite{Matterport3D}. 
% From the experimental study, our navigation framework capitalizes on the SEA to achieve a 39.0\% success rate. This result represents a significant 12.4\% improvement over the current state-of-the-art, underlining the SEA's potential to reshape the landscape of embodied agent navigation. 

%------------------------------------------------------------------------
\section{Related Work}
\paragraph{Visual navigation without any map} 
As a policy network, a recurrent neural network (RNN) is a simple method to make an implicit semantic prior~\cite{DDPPO, RedRabbit, THDA}. 
DDPPO~\cite{DDPPO} has a vanilla RL policy with a CNN backbone followed by an LSTM as a policy function.
Red-Rabbit~\cite{RedRabbit} augments DDPPO with multiple auxiliary tasks, such as predicting agent dynamics, environment states, and map coverage with ObjectNav.
Treasure Hunt Data Augmentation (THDA)~\cite{THDA} improves the RL reward and model inputs, which result in better generalization to new scenes. %It artificially inserts objects while training.
Since an RNN has a difficulty of backpropagating a long sequence, an RNN can be replaced with an explicit structure~\cite{ANS, SemExp, PONI, FBE, Online, OTZ, bayesian, HRE}. 

\paragraph{Visual navigation with a metric map} 
Spatial metric map-based RL methods~\cite{ANS, SemExp, PONI, FBE} propose independent modules for semantic mapping, high-level semantic exploration, and low-level navigation. 
The semantic exploration module is learned through RL, yet it is more sample-efficient and generalizes better than end-to-end RL. 
Active Neural SLAM (ANS)~\cite{ANS} has a hierarchical structure to explore an environment: global and local policies. 
The global policy constructs a top-down 2D map and estimates a global goal.
Given the global goal from the global policy module, a local policy module plans a path to the goal using a simple local navigation algorithm.
Semantic exploration~\cite{SemExp} is a study that extends ANS. 
The metric map does not only represent obstacles but draws a semantic map and uses it for navigation to improve performance. 
This method implicitly learns semantic information for navigation.
PONI~\cite{PONI} reduced computational costs in visual navigation by proposing non-interactive learning.
Additionally, it improved the navigation performance by learning the encoder by calculating the probability that there is a space or an object beyond the frontier boundary of the current map and then moving to the boundary where the object is likely placed.
However, since this method is trained using a top-down map, it is greatly affected by the pose sensor.
% 
% It has the disadvantage of being vulnerable to a situation where two objects of different categories overlap.

\paragraph{Visual navigation with a graph map} 
Our work proposes a method to collect semantic priors and use it for navigation.
Several works have employed semantic priors into a graph to enhance semantic reasoning in visual navigation~\cite{Online, OTZ, bayesian}.
Wu et al.~\cite{bayesian} tackle the room navigation task using room relationship, while it does not consider the relationship between a room and an object.
Zhang et al.~\cite{OTZ} divide a room into several zones to find an object and find the reachability between these zones for navigation. 
However, the connection between an object and a zone is ambiguous. 
For example, a bed can exist in any zone in a bedroom.
% 
% Therefore, it cannot be said that the relationship between an exact place and an object has been learned.
% 
Campari et al.~\cite{Online} improve performance by building an abstract model in addition to the existing metric map-based methods. 
Here, the abstract model comprises nodes composed of images and objects, and the connection between nodes is an action taken to navigate between two places.
However, actions for moving from one place to another could differ depending on the structure of houses.
For example, in one house, the bedroom may be to the right of the living room, and in another, the bedroom may be to the left. 
Therefore, the structure of the environment is hard to be expressed with the abstract model.

The proposed method collects relationships between place clusters and objects using a sequence of observations and uses them in a new environment for navigation. 
% 
% Using the place and object relationship, the target place can be estimated. 
% 
% Additionally, semantic path planning is possible using the place relationship.

\begin{figure*}[t!]{\includegraphics[width=\textwidth]{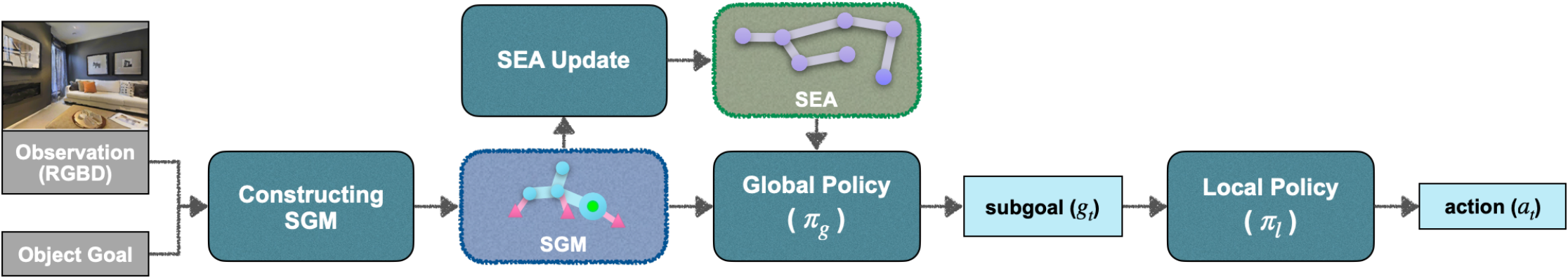}}
\caption{
\textbf{Overview of semantic environmental atlas (SEA).} 
The semantic graph map (SGM) is updated using visual observations. %: an episodic graph, semantic prior graph, global policy, and local policy.
% 
% An episodic graph encodes semantic relationship using the sequence of observations.
% 
Then, the place relationship and place-object connections across environments are updated using multiple semantic graph maps collected from different environments.
A global policy samples a subgoal $g_t$, which is reachable and most likely to be near to the target place.
A local policy generates navigational actions to reach the subgoal. 
}
% See text for more details.
\label{fig:overview}
\end{figure*}
\section{Proposed Method}
\subsection{Problem Statement}\label{sec:section3.1}
In a given unknown environment, an agent is tasked with traveling to an object specified by its category name (e.g., chair) (see Figure\ref{fig:overview}). 
At the start of each episode ($t=0$), the agent is placed at a random navigable position within the environment. The agent is equipped with a 640 $\times$ 480 RGB-D sensor ($s^d_t$) and a 512 $\times$ 128 panoramic RGB sensor ($s^p_t$), along with the goal category ($O_{\text{goal}}$) for the current time step. The panoramic RGB sensor ($s^p_t$) is specifically used for constructing the semantic graph map. It is important to note that a pose sensor is employed only in the local policy, and global pose sensor readings are not used in this work. The agent can perform actions $a_t$ $\sim$ $\mathcal{A}$, where $\mathcal{A}$ includes moving forward (0.4m), turning left (30$\degree$), turning right (30$\degree$), and stopping. To complete the task, the agent must press the stop button once it is within a success distance of $d_s$ = 1.0m from the target. The episode concludes either when the agent stops or when the time budget of $T$ = 500 steps is exceeded.
% At the beginning of each episode ($t=0$), the agent is placed at a random navigable position within the environment. The agent is provided with a 640 $\times$ 480 RGB-D sensor value $s^d_t$ and a 512 $\times$ 128 panoramic RGB sensor value $s^p_t$, as well as the goal category $O_{\text{goal}}$ at time $t$. The panoramic sensor $s^p_t$ is only used for constructing the semantic graph map. It is important to note that a pose sensor is only used in the local policy and global pose sensor readings are not utilized in this work. The agent then performs an action $a_t$ $\sim$ $\mathcal{A}$, where $\mathcal{A}$ consists of move forward (0.4m), turn left (30$\degree$), turn right (30$\degree$), and stop. In order to complete the task, the agent must press the stop button when it reaches within $d_s$ $=$ 1.0m of the target. The episode ends when the agent performs a stop or when the time budget of $T$ $=$ 500 steps is exceeded.

% The task of object goal navigation is divided into two parts: finding the location of the object and navigating to that location. This problem is tackled by defining place-object and place relationships. Priors are collected from semantic graph maps, which consist of images and associated objects, without the use of a pose sensor.

\subsection{Semantic Graph Map}
\begin{figure*}[t!]{\centering\includegraphics[width=\textwidth]{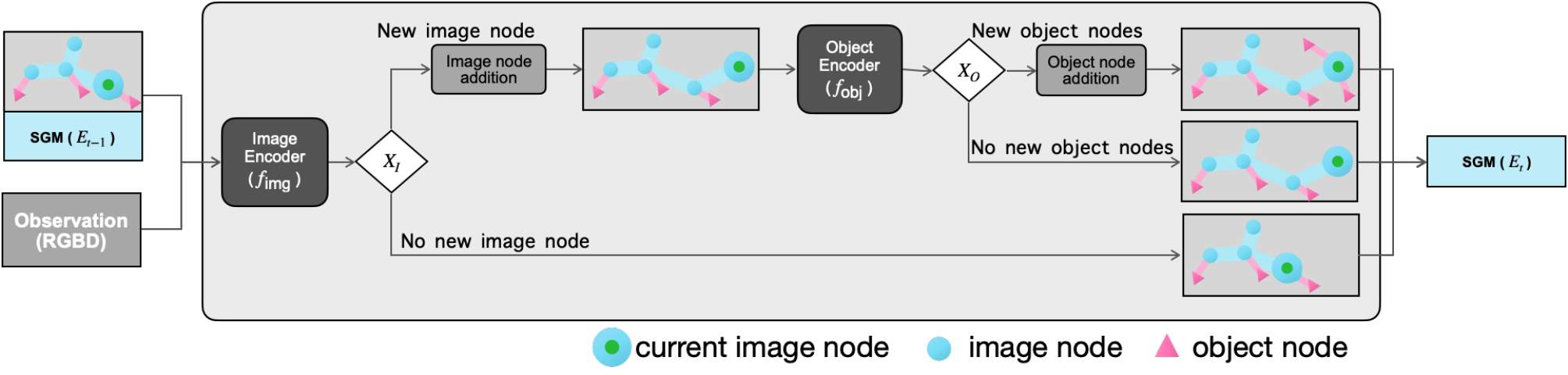}}\centering\squeezelittle
\caption{
\textbf{Construction of semantic graph map.}
By integrating the current observation and the previous semantic graph map (SGM; $E_{t-1}$), the graph map is updated. 
If it is discovered that the current location differs from the previous location, an image node is added to the graph. 
Similarly, object nodes are added to the graph when previously undetected objects are detected.
}
\label{fig:episodicgraph}
\end{figure*}

Inspired by the topological graph map approach outlined in \cite{TSGM}, we construct semantic graph maps, $E_t$, for navigation in unknown environments, as depicted in Figure~\ref{fig:episodicgraph}. At each time point $t$, the semantic graph map includes three types of nodes—place nodes ($\mathcal{V}_{\text{place}}$), image nodes ($\mathcal{V}_{\text{im}}$), and object nodes ($\mathcal{V}_{\text{ob}}$)—and corresponding edges: $\mathcal{E}_{\text{im}}$, $\mathcal{E}_{\text{io}}$, and $\mathcal{E}_{\text{pi}}$. Each place node, represented as ${P_i}$, connects to image nodes with an affinity matrix $\mathbf{A}_{\text{im}} \in \mathbb{R}^{N_i \times N_i}$ indicating the relational strengths. The object nodes, ${x_i}$ where $x_i \in \mathbb{R}^{1 \times D_o}$, are similarly linked to image nodes with an affinity matrix $\mathbf{A}_{\text{io}} \in \mathbb{R}^{N_i \times N_o}$. The edges $\mathcal{E}_{\text{pi}}$ connect place and image nodes with an affinity matrix $\mathbf{A}_{\text{pi}} \in \mathbb{R}^{N_p \times N_i}$, illustrating the relationships between places and images.
The affinity matrices are computed using a multi-layer perceptron (MLP) network, which processes the features of nodes to output a scalar similarity value. The semantic graph map is constructed incrementally as the agent navigates, with the graph at time $t$ being a subset of the graph at time $t+1$. This dynamic mapping allows the agent to reason about and navigate through the relationships among objects, images, and places towards the designated goal.

\paragraph{Place graph} \label{sec:place_encoder}
In visual navigation, accurately identifying semantic places, such as living rooms and bedrooms, is crucial. To address this challenge, room navigation methods~\cite{bayesian, unscene} employ a place recognition algorithm. However, some places can be ambiguous and difficult to classify distinctly. 
% In visual navigation, finding semantic places, such as living rooms and bedrooms, is beneficial. To tackle this problem, room navigation methods~\cite{bayesian, unscene} employ a place recognition algorithm. However, there could be ambiguous places that are difficult to identify as a single place. 
To overcome this issue, a clustering method~\cite{PCL} is applied to train a place encoder, $f_{\text{place}}$, which groups similar features across similar places. This encoder takes as input image features, object features, and object categories to extract place information, defined as $v_t = f_{\text{place}}(s_t^p, o^\text{f}_i, o^{\text{cat}}_i)$. Here, $o_i$ represents objects detected from a panoramic RGB sensor  $s_t^p$; $o^\text{f}_i$ is a feature vector of $o_i$; and $o^{\text{cat}}_i$ is the object's category. 
Using a panoramic RGB sensor is advantageous because the recognition of the place is invariant to camera rotation. To bring images with similar semantic meanings, such as those from a bedroom, closer together in the metric space, a contrastive loss is employed. The loss function for training the place encoder with a batch of $B$ images is formulated as follows:
\begin{align}
\begin{aligned}{
\mathcal{L}_{\textit{Place}} = \sum_{i=1}^{B} -\log\frac{\exp{(v_i \cdot v_i^{'} / \zeta)}}{\sum_{j=0}^{r}\exp{(v_i \cdot v_j^{'} / \zeta)}},
}\end{aligned}
\end{align}
where $v_i$ is the query embedding and $v_i^{'}$ are positive place embeddings for location $i$, sampled from the same place, and $v_j^{'}$ includes one positive embedding and $r$ negative embeddings from different places, with $\zeta$ acting as a temperature hyper-parameter. 
Positive samples are drawn using ground truth place information which includes labels for ambiguous places such as `other room.' To handle this, we utilize eight specific room labels, described in Section~\ref{experimental-settings}. We modify the contrastive loss by replacing the positive sample with images generated by randomly rotating the query image, a method we denote as $\mathcal{L}_{\text{Near}}$. This approach aims to bring images from nearby locations closer together in the metric space. Subsequently, we apply the $K$-means clustering algorithm to cluster features, resulting in a set $\mathbb{P} = \{P_1, ..., P_{N_p}\}$. These clustering results are then used as the ground truth for the clustering loss. Metric learning involves iteratively combining all losses, $\mathcal{L}_{met} = \mathcal{L}_{\text{place}} + \mathcal{L}_{\text{Near}} + \mathcal{L}_{\text{Cluster}}$, with the $K$-means clustering process.

% start revising from here
\paragraph{Image graph}
An image encoder~\cite{VGM}, represented as $i_t = f_{\text{img}}(s^p_t)$, is crucial for assessing image similarity to determine the novelty of nodes in the semantic graph map. When the agent moves to a new location, it evaluates the similarity between the current and previous image nodes using a cosine similarity function, $\text{sim}(\cdot,\cdot)$. If this similarity, $\text{sim}(i_{t-1}, i_{t})$, drops below the threshold $\theta_{\text{im}} = 0.8$, the system checks if the observed image node already exists in the graph. If not, indicating no similar existing image nodes, the node is considered new ($i_t$) and connected to the previous image node ($i_{t-1}$). Conversely, if a similar node is found, it is updated with the new image and also linked to $i_{t-1}$. Image graph construction allows the system to capture and represent the spatial relationships between different places and objects within an environment. This spatial representation is crucial for tasks that require an understanding of the layout of an environment, such as navigation and path planning. The adjacency matrix of image nodes, $\mathbf{A}_{\text{im}} \in \mathbb{R}^{N_i \times N_i}$, is a binary matrix that records these connections. Additionally, when a new image node is detected, it and the place cluster associated with the image node are linked, setting $\mathbf{A}_{\text{pi}}[i, j] = 1$ for the $i$th place and $j$th image, reinforcing the semantic links.

\paragraph{Object graph}
Objects are encoded using an object encoder, $x_i = f_{\text{obj}}(s^p_t, o^\text{f}_i, o^{\text{cat}}_i)$, which uses contrastive learning to recognize objects as the same even when viewed from different perspectives~\cite{TSGM}. Here, $o_i$ represents an object detected from $s^p_t$ using MaskRCNN~\cite{he2017mask}. The graph update module assesses whether these objects are already present in the graph. If the similarity between the detected object and an existing object in the graph, $\text{sim}(x_i, x_j)$ is greater than $\theta_{\text{o}} = 0.8$ and their categories match ($o^{\text{cat}}_i = o^{\text{cat}}_j$), the object is considered the same. If the detected object is not in the graph, it is added as a new object node and linked to the current image node ($i_t$), with $\mathbf{A}_{\text{io}}[i, j] = 1$ indicating the connection between the $i$th image node and the $j$th object node. Conversely, if the object is already in the graph but the detected object has a higher detection score, the existing object node is updated to reflect the new detection.

\subsection{Localization}
In this study, we test the localization function, ${F}_{loc}$, to demonstrate how semantic knowledge can enhance localization accuracy. The input to this function includes the current semantic graph map, $\mathcal{G}_t$, along with source and target images. We employ Graph Neural Networks (GNNs) to encode the graph data effectively and Transformer~\cite{vaswani2017attention} decoder networks to extract relevant localization information. Specifically, ${F}_{loc}$ utilizes these images to identify corresponding nodes within the graph. It then estimates the distance between these nodes, thereby facilitating precise localization based on semantic relationships captured in the graph.

\begin{figure*}[h!]{\centering\includegraphics[width=\textwidth]{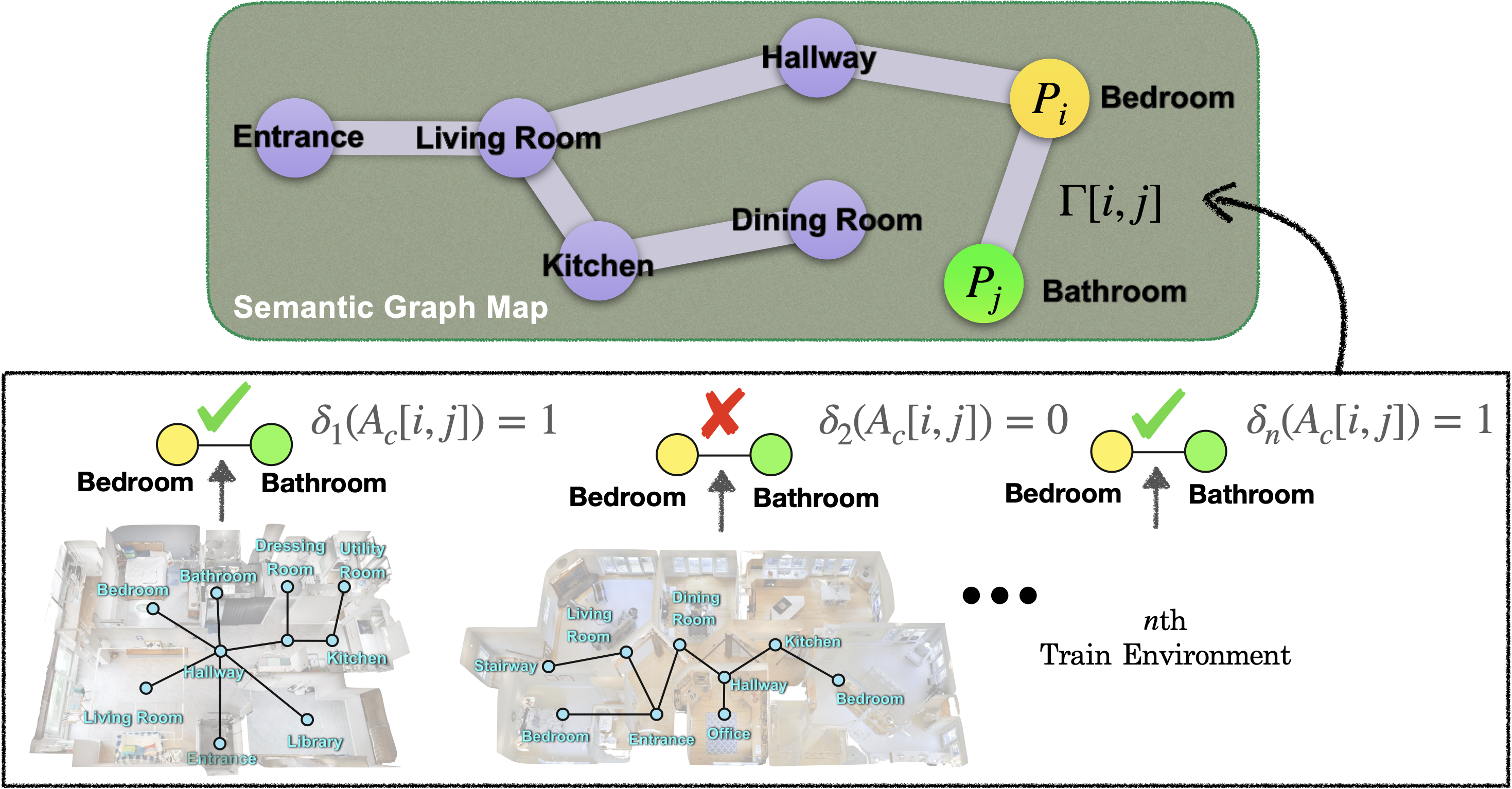}}\centering
\caption{
\textbf{Formation of Semantic Graph Map from Training Environments.} The figure illustrates the process of forming a semantic graph map from multiple training environments.
In each training environment, it checks to see whether there is a pair of place clusters.
If there a connection between the pair of places, the reachability is set to one; otherwise, zero.
}
\label{fig:place_connection}
\end{figure*}

% \subsection{Graph Integration} 
\subsection{Semantic Environment Atlas} 
% Graph intergration module? semantic atlas?
We propose a Semantic Environment Atlas (SEA) that synthesizes semantic graph maps collected from various environments into a unified structure, rather than merely compiling individual maps. This integration facilitates a deeper and more comprehensive understanding of the environments and their interrelationships. The SEA, denoted as $S_t = \{\mathbf{\Gamma}, \mathbf{R}\}$, comprises two key components: a place reachability matrix ($\mathbf{\Gamma}$) and a place-object connection matrix ($\mathbf{R}$). he place reachability matrix ($\mathbf{\Gamma}$) defines the accessibility between different places, indicating possible paths and their navigability. Meanwhile, the place-object connection matrix ($\mathbf{R}$) details the associations between various places and the objects found within them, providing crucial contextual information that enhances navigational decisions and spatial reasoning.

% Combining the semantic graph maps collected from different environments, we suggest a semantic atlas that is not just creating individual maps, but is organizing and integrating these maps in a way that provides a more complete understanding of the environments and their relationships to each other.
% The SEA ($S_t = \{\mathbf{\Gamma}, \mathbf{R}\}$) is composed of a place reachability matrix ($\mathbf{\Gamma}$) and a place-object connection matrix ($\mathbf{R}$).

\paragraph{Place-place relationship}
A node within the place graph represents the centroid of a place cluster, and connections between nodes signify the reachability between places. This reachability is derived from semantic graph maps collected across various training environments. If two connected image nodes, $i_a$ and $i_b$, belong to different place clusters, $P_a$ and $P_b$, it is inferred that the clusters are reachable. A connection between clusters in any scene sets the cluster reachability to one; otherwise, it is zero. This procedure is repeated across all training scenes, and the average value is taken as the final measure of reachability between places. 
The formation of the semantic graph map from episodic graphs in training environments can be shown in Figure~\ref{fig:place_connection}. The top section of the figure shows a semantic graph map where nodes represent different rooms (e.g., Entrance, Living Room, Kitchen) and edges represent the connectivity between them. For example, the edge \( \Gamma[i,j] \) connects nodes \( P_i \) (Bedroom) and \( P_j \) (Bathroom), indicating a valid connection in the graph.
In the bottom section of the figure, $n$ floor plan layouts are depicted, each demonstrating the connectivity between rooms in various training environments. In the first floor plan, the connection between the bedroom and bathroom is correctly identified (\( \delta_1(A_c[i,j]) = 1 \)), as indicated by the green check mark. In the second training environment, the connectivity between bedroom and bathroom is not recognized (\( \delta_2(A_c[i,j]) = 0 \)), marked by the red cross. When considering  $N$ training environments, the reachability ($\mathbf{\Gamma}$) between place cluster $i$ and $j$ is calculated as follows:
% , it is assumed that the place clusters $P_a$ and $P_b$ are reachable.
% If a cluster connection appears in a scene, the cluster reachability is set to one; otherwise, it is set to zero.
% We repeat this process for all training scenes and take the mean value as the final reachability between places.
% When there are $N$ training environments, reachability ($\mathbf{\Gamma}$) between place cluster $i$ and $j$ can be represented as:
\begin{align}
\begin{aligned}
\mathbf{\Gamma}[i, j] = \frac{\sum^N_{n=1}\delta_n(\mathbf{A}_\text{c}[i, j])}{\sum^N_{n=1}\delta_n(P_i)\delta_n(P_j)^T},
\end{aligned}
\end{align}
where $\mathbf{A}_\text{c} = \mathbf{A}_{\text{pi}}\mathbf{A}_{\text{im}}\mathbf{A}_{\text{pi}}^T$ and $\delta_n(\mathbf{A}_\text{c}[i, j])$ indicates the existence of a connection between place cluster $i$ and $j$. The function $\delta_n(P_{i})$ denotes the presence of the place cluster $P_{i}$ in the $n$th scene.

% is the existence of the connection between place cluster $i$ and $j$, and $\delta_n(P_{i})$ is the existence of the place cluster $P_i$ on the $n$th scene.
It is important to note that the same places are not connected, thus $\mathbf{A}_\text{c}[i, j] = 0$ when $i = j$.
Furthermore, to normalize the reachability, we calculate it based on the number of environments in which each cluster appears, rather than the total number of environments. Reachability is set to zero if a cluster does not appear in any scene.

\paragraph{Place-object relationship} \label{sec:place_to_object}
The relationship between object nodes and place clusters is established by connecting object nodes to image nodes within a graph, and then linking these image nodes to place graph nodes. This linkage facilitates the computation of the probability distribution for place and object categories. Specifically, for the $n$th training environment, we calculate $\mathbf{A}^n_{\text{po}} = \mathbf{A}^n_{\text{pi}} \mathbf{A}^n_{\text{io}} \mathbf{A}^n_{\text{oc}}$, where $\mathbf{A}_{\text{oc}} \in \mathbb{R}^{N_o \times N_c}$ maps each object node to its corresponding object category and $N_c$ is the number of object categories.

By aggregating all semantic graph maps from the training environments, the relational connection between each place cluster and the object categories is defined as $\mathbf{R} = \sum_{n=1}^N \mathbf{A}^n_{\text{po}}$, where $\mathbf{R} \in \mathbb{R}^{N_p \times N_c}$ represents the number of connections between place clusters and object categories.

Given a set of object categories $\mathbb{O} = \{{O}_1,..., {O}_{N_c}\}$, the probabilities of encountering a specific place cluster $i$ given an object category $j$, and conversely, the probability of encountering an object category $j$ given a place cluster $i$, are computed as follows:
\begin{align}
\begin{aligned}
p(P_i|{O}_j) = \frac{\mathbf{R}[i, j]}{\sum_{k=1}^{N_p} \mathbf{R}[k, j]}, 
p(O_j|P_i) = \frac{\mathbf{R}[i, j]}{\sum_{c=1}^{N_c} \mathbf{R}[i, c]},
\end{aligned}
\end{align}
where $\mathbf{R}[i, j]$ indicates the number of connections between place cluster $i$ and object category $j$. These probability distributions are illustrated in Section 2 of the supplementary material. 

\begin{figure*}[t!]
\centering
\begin{minipage}{\textwidth}{
\centering
{\centering\includegraphics[width=\textwidth]{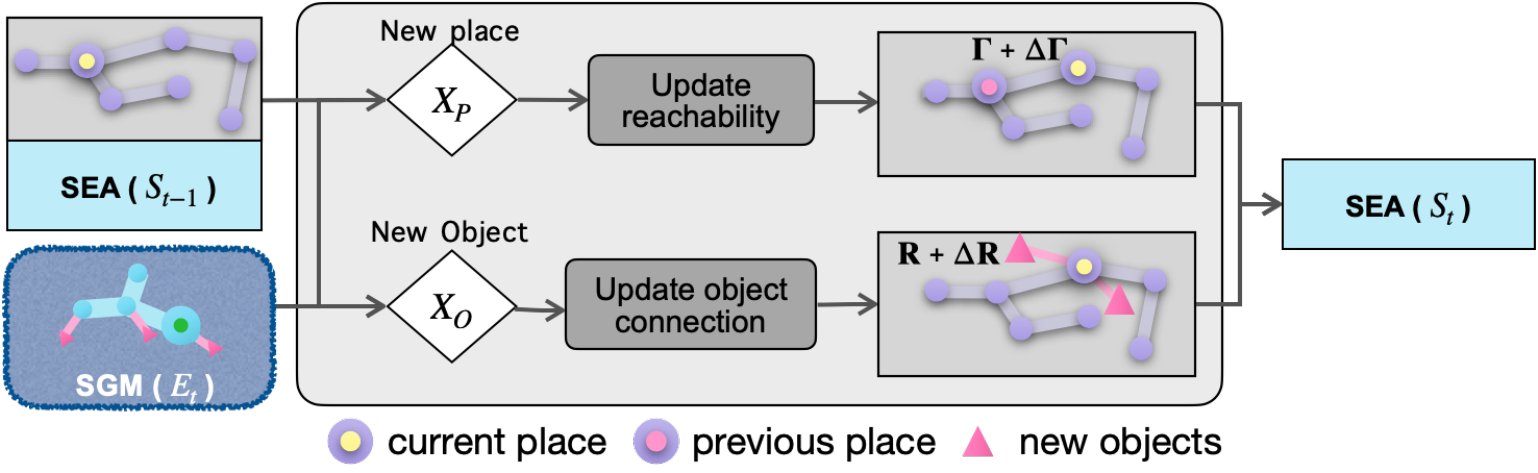}}
\centering
\caption{
{Adaptive SEA update procedure.}
% 
% The semantic prior graph and episodic graph are used to update the prior information. 
% 
% The semantic prior graph is updated by extracting place-object and place connections from episodic graphs.
% 
}
\label{fig:SEA}}
\end{minipage}
\end{figure*}

\paragraph{Updating relations}
Our experimental setup is designed to test the navigation agent's ability to plan its path using common sense, akin to human navigation, in a new environment without a pre-existing map. To adapt effectively to these unfamiliar settings, the SEA updates the graph in a Bayesian manner~\cite{bayesian}. As illustrated in Figure~\ref{fig:SEA}, when a new place cluster or object-place connection is discovered during the construction of the semantic graph map, the prior probability is adjusted based on the new observations to calculate the posterior distribution. Given that these probabilities are determined by counting occurrences, the impact of new connections is generally minor. Thus, the update rate is set at 0.1 of the maximum count value ($\max(\mathbf{R}[i, j])$). The graph is continuously updated at every step. If the target object is not detected in the expected target place, the probability associated with the target object being in that place decreases. As the connection between the target object and the place cluster weakens, the next most connected place cluster is identified and explored.

% The test environment may differ from the training scenes. To address this, the SEA is updated in a Bayesian manner~\cite{bayesian}.
% As seen in Figure~\ref{fig:SEA}, if a new place cluster or object-place connection is discovered while constructing the semantic graph map, the prior probability is adjusted using observations to obtain the posterior distribution.
% Since the probability is defined by counting, the impact of new connections is likely to be minor. As such, the update rate is set to 0.1 of the maximum count value ($\max(\mathbf{R}[i, j])$).
% The graph is updated at every step, and if the target object is not detected in the target place, the probability of the target object and the target place decreases. As the connection between the target object and the place cluster weakens, the place cluster with the next highest connection is subsequently discovered.

\subsection{Global Policy}
{\color{black}
The global policy ($\pi_g$) utilizes the RGB-D image from the directional camera ($s^d_t$) to determine subgoals ($g_t$) through semantic path planning, which leverages the place relationships and place-object relationships within the SEA.
% Using this information, the agent computes the shortest semantic path between locations within the environment. 
% 
For example, to navigate from a bathroom to a kitchen, the calculated path might be bathroom $\rightarrow$ bedroom $\rightarrow$ living room $\rightarrow$ kitchen. Rather than attempting to locate the kitchen directly from the bathroom, the agent first navigates to intermediary nodes such as the bedroom and the living room, thereby systematically discovering the optimal path. 
% By setting intermediary subgoals based on the SEA, the agent ensures efficient and accurate navigation.
}
% The global policy ($\pi_g$) utilizes the RGB-D image from the directional camera ($s^d_t$) to determine subgoals ($g_t$), which are identified through semantic path planning. This planning leverages the place relationship and place-object relationship within the SEA.

\paragraph{Current place}
{\color{black}
To enable the semantic path planning, the agent first determines its current location by encoding RGB images and object information using the place encoder (as detailed in Section~\ref{sec:place_encoder}).
The extracted place feature, derived from the encoded RGB image, object features, and object category, is then compared to place clusters ($P$) using cosine similarity to identify the nearest place cluster, thereby locating the agent within the environment.
}
% To ascertain the current location within the environment, the current place feature is extracted by encoding the RGB image, object features, and object category using the place encoder. The encoded place feature is then compared to the place clusters ($P$) using cosine similarity to identify the nearest place cluster, effectively locating the agent within the map.
% The current place feature can be obtained by encoding the RGB image, object features, and object category in the place encoder (as defined in Section~\ref{sec:place_encoder}). By comparing the place feature to the place clusters $P$ using cosine similarity, the closest place cluster can be identified.

\paragraph{Target place}
The target place is chosen based on its association with the target object. The place cluster with the highest probability of containing the target object is selected as the target destination. This selection process is mathematically formulated as follows:
% The place cluster with the highest probability given the target object is selected as the target place. Mathematically, this can be described as:
\begin{align}
\begin{aligned}
    k^*_t = \text{argmax}_{k} p_t(P_k|O_{\text{goal}}),
\end{aligned}
\end{align}
where $P_{k}$ represents the potential target place, and $p_t(P_k|O_{\text{goal}})$ denotes the probability of place cluster $k$ given the target object $O_{\text{goal}}$, as defined in Section~\ref{sec:place_to_object}.

\paragraph{Subgoal place}
Using conventional graph-based planning methods, a navigation agent can identify an optimal trajectory to the target place. However, the optimal subgoal may not always be near the current location. To address this, subgoal candidates are selected among the visible places identified using detected objects from the directional sensor.

% Using conventional graph-based planning methods, a navigation agent can find an optimal trajectory to the target place. 
% The place relationships
% with conventional graph-based planning methods, such as ~\cite{}.
% However, the optimal subgoal may not be located near the current place.
% Therefore, subgoal candidates are selected among the visible places, using detected objects from the directional sensor.
To streamline the selection process and reduce computational complexity, the probability of each place, $p_t(\mathbb{P}|o_t)$, is approximated using the importance of the object category.
The object importance is defined as the inverse of the entropy of the object distribution conditioned on places, given as $1/\mathbb {E}_\mathbb{P}[-\log p_t(\mathbb{O}|\mathbb{P})]$.
Objects associated with a single place cluster have high importance, whereas those common to multiple clusters exhibit lower importance.

Among the observed objects, the category deemed most important, denoted $o$, is used to determine the place cluster with the highest probability, calculated as $\text{argmax}_\mathbb{P} p(\mathbb{P}|o)$.
To facilitate this, the directional image with a field of view (FOV) of 120$\degree$ is narrowed by 40$\degree$ to focus on objects directly ahead ($o_{f*}$), to the left ($o_{l*}$), and to the right ($o_{r*}$).
These selected objects help estimate the subgoal candidates: $\mathbb{P}_s=\{P_{f}, P_{l}, P_{r}\}$, where $P_x = \text{argmax}_{\mathbb{P}} p_t(\mathbb{P}|o_{x*})$ for $x$ representing the front, left, and right directions, respectively.

A subgoal place ($g_t$) with the highest reachability to the target place cluster is then chosen from these subgoal places. The selection is based on the following formula:
\begin{align}
\begin{aligned}
    g_t = \text{argmax}_{P_{\tau_0} \in \mathbb{P}_s} \prod_{i=1}^{m-1}{\mathbf{\Gamma}_{P_{\tau_{i-1}}P_{\tau_{i}}} \cdot \mathbf{\Gamma}_{P_{\tau_{m-1}}{P_{k^*_t}}}},
\end{aligned}
\end{align}
where $\{P_{\tau_0}, ..., P_{\tau_{m-1}}, P_{k^*_t}\}$ represents the optimal semantic path from the subgoal to the target place.
If a subgoal is beyond a reasonable straight-line distance, it is considered unreachable and is excluded, similar to the NRNS method~\cite{NRNS}.
% 
% Subsequently, the candidate that is both reachable and closest to the selected semantic subgoal is chosen as the subgoal.
The remaining candidate that is both reachable and closest to the semantic subgoal is selected as the final subgoal.
If all potential subgoals belong to the same place cluster, or if no detected objects aid the decision, a subgoal is randomly chosen among them. This mechanism encourages broader exploration by the agent, preventing it from being confined to a specific area.
% Furthermore, if the place clusters to which subgoals belong are all the same, or if no detected objects are found, a subgoal is chosen randomly between them.
% 
% This process facilitates greater exploration by an agent, thereby preventing it from getting stuck in a particular region.

Furthermore, the semantic path is derived from a shortest path calculation by setting the edge weight in the place graph between $i$th place and $j$th place to $-\log(\mathbf{\Gamma}_{P_i P_j})$:
\begin{align}
\begin{aligned}
    \Tau^* = \text{argmin}_{\tau} \exp \sum_{i=1}^m -\log(\mathbf{\Gamma}_{P_{\tau_{i-1}}P_{\tau_{i}}}),
\end{aligned}
\end{align}
where $\Tau^* = \{\tau_0, ..., \tau_m\}$ is a set of indices representing the optimal semantic path, starting from $P_{\tau_0}$ and ending at $P_{\tau_m}$, the goal place. 
This trajectory, $\Tau^*$, represents the most probable path, effectively bridging the start and the target locations, optimizing the agent's navigation strategy.

\subsection{Local Policy}
The local policy ($\pi_l$) processes directional RGB-D sensor data ($s^d_t$) along with local pose sensor readings to navigate the agent towards the designated subgoal $g_t$.
It employs the fast marching method (FMM)~\cite{FMM} to compute the shortest path from the agent's current location to the subgoal. This computation makes use of the obstacle channel, which is derived from the top-down map created from the depth component of the RGB-D input.
%
% Once the shortest path is determined, the local policy makes deterministic actions to navigate the agent along the path.
%
% Previous research has shown this approach to be successful~\cite{ANS, SemExp, PONI}.
Upon determining the shortest path, the local policy executes a series of deterministic actions to guide the agent along this path. This strategy of navigation has been validated in previous research, demonstrating its effectiveness in various scenarios~\cite{ANS, SemExp, PONI}.

\section{Experiments}
\subsection{Baselines}
\paragraph{Non-interactive baselines}
\textbf{\texttt{BC}}: A baseline for behavior cloning was trained using an RNN-based policy that takes RGB-D, agent pose, and goal object category as inputs.
\paragraph{End-to-end RL baselines}
\noindent\textbf{\texttt{DD-PPO}\cite{DDPPO}}: Standard end-to-end RL with distributed training over several nodes is proposed.
\noindent\textbf{\texttt{Red-Rabbit}\cite{RedRabbit}}: Auxiliary tasks that improve sampling efficiency and generalization to previously unseen domains are provided.
\noindent\textbf{\texttt{THDA}~\cite{THDA}}: RL reward and model inputs are improved, which results in better generalization to new scenes.
\paragraph{Metric map-based baselines}
\noindent\textbf{\texttt{FBE}\cite{FBE}}: A traditional frontier-based exploration method is adapted to object goal navigation using a detector to detect the target.
It triggers a stop when the target is reached using the metric map.
\noindent\textbf{\texttt{ANS}\cite{ANS}}: A spatial metric map-based RL policy trained for exploration is adapted to the object goal navigation using the same heuristic as \textbf{\texttt{FBE}}\cite{FBE} for goal detection and stopping.
\noindent\textbf{\texttt{PONI}\cite{PONI}}: Non-interactive training is used to navigate and only trained potential fields are used to determine the next subgoal.
\paragraph{Graph map-based baselines}
\noindent\textbf{\texttt{ANS + SI}\cite{Online}}: An abstract model is attached to \textbf{\texttt{ANS}\cite{ANS}}. The agent incrementally extends the abstract model and reuses the learned model from previous episodes. 
\noindent\textbf{\texttt{SemExp + SI}\cite{Online}}: An abstract model is attached to semantic exploration (\textbf{\texttt{SemExp}\cite{SemExp}}) using the same abstract model strategy as \textbf{\texttt{ANS + SI}}.
\noindent For \textbf{\texttt{DD-PPO}}, \textbf{\texttt{Red-Rabbit}}, \textbf{\texttt{THDA}}, and \textbf{\texttt{PONI}}, publicly available MP3D results on the Habitat ObjectNav leaderboard are used.
For \textbf{\texttt{ANS}}, pre-trained models released by the authors are evaluated.
For \textbf{\texttt{ANS + SI}} and \textbf{\texttt{SemExp + SI}}, official results from the published paper are used.

\subsection{Experimental Settings} \label{experimental-settings}
\paragraph{Datasets}
We utilized the Habitat simulator~\cite{habitat19iccv} to conduct experiments using the Matterport3D (MP3D)\cite{Matterport3D} datasets, which feature photorealistic 3D reconstructions of the real world.
The standard 61 train / 11 val splits for the ObjectNav configuration, as described in Section\ref{sec:section3.1}, were employed.
It should be noted that only the local policy depends on the depth and pose, making the proposed method considerably more practical for use in the real world with noisy pose and depth sensors.
The Habitat ObjectNav dataset~\cite{habitat19iccv} was used for MP3D experiments, with 21 goal categories (provided in the supplementary Section 1). 
% The MP3D~\cite{Matterport3D} encompasses 40 object categories, while the ObjectNav dataset from the Habitat challenge contains 21 specific object categories as goals. 
% These categories are as follows:
% \textit{chair}, \textit{table}, \textit{picture}, \textit{cabinet}, \textit{cushion}, \textit{sofa}, \textit{bed}, \textit{chest of drawers}, \textit{plant}, \textit{sink}, \textit{toilet}, \textit{stool}, \textit{towel}, \textit{tv monitor}, \textit{shower}, \textit{bathtub}, \textit{counter}, \textit{fireplace}, \textit{gym equipment}, \textit{seating}, and \textit{clothes}. 
% Considering some object categories in MP3D are not applicable for place-object statistics, we decided to utilize 32 specific object categories, as illustrated in Figure~\ref{fig:object_importance}. The following eight categories were excluded: \textit{wall}, \textit{floor}, \textit{ceiling}, \textit{column}, \textit{lighting}, \textit{beam}, \textit{railing}, \textit{misc}.
% Similarly, for places, MP3D comprises 30 ground truth places. However, due to the ambiguity of places such as \textit{other room} and \textit{junk}, we opted for eight evident places for place clustering: \textit{living room}, \textit{bedroom}, \textit{kitchen}, \textit{closet}, \textit{dining room}, \textit{bathroom}, \textit{toilet}, \textit{hallway}.
%
% No training episodes from the ObjectNav dataset~\cite{habitat19iccv} were used, 
We utilized 2195 episodes for the test.

\paragraph{Evaluation metrics}
All methods were evaluated using the success rate (\textbf{Success}), success weighted by path length (\textbf{SPL})~\cite{SPL}, and distance to success (\textbf{DTS}).
\textbf{Success} is determined by calculating the ratio of successful test episodes to the total number of test episodes.
\textbf{SPL} takes into account both the Success and path length.
When there are $M$ episodes, $\textbf{SPL} = \frac{1}{M} \sum_{i=1}^M Y_i \frac{l_i}{\max(p_i, l_i)}$, where $l_i$ is the length of the shortest path from goal to target, $p_i$ is the length of the path taken by the agent, and $Y_i$ is the binary indicator of Success for $i$th episode.
Finally, \textbf{DTS} is the $L_2$ distance (measured in $m$) between the agent and the success threshold (1.0$m$) of the goal object at the end of the episode, as described in ~\cite{SPL}.

\paragraph{Implementation details}
To construct SEA, we examined ten episodes from each train scene, for a total of 610 episodes.
The maximum number of time steps was set to 500, and the environment was explored randomly.
For object detection, we trained a MaskRCNN~\cite{he2017mask} model to identify 40 object categories in MP3D environments.
Our method does not construct a metric map, thus a different stopping criterion was used compared to metric map-based methods.
If an object is detected from a distance and exceeds a target object detection score threshold, the agent approaches the object and checks whether it is the target object.
If the object detection score is lower after the encounter, it is assumed that it is not the target.
The object feature with the highest detection score along the approaching path is stored in the checked object list.
If a detected target object is highly similar to the checked objects, it is not rechecked, as it has already been searched.
Additional details can be found in Section 1 of the supplementary material.
% Please let me know if there is anything else I can help you with.

\subsection{Results}

\begin{table}[!t]
\centering
% \begin{minipage}{.5\textwidth}{
\centering
\caption{{Habitat ObjectNav results on MP3D. We report the
results from the top-performing methods. %${}^\dag$ This is privileged.
}}
\resizebox{\linewidth}{!}{
\begin{tabular}{lc|ccc} \toprule 
\multirow[b]{2}{*}{Method} & \multirow[b]{2}{*}{Pose noise} & \multicolumn{3}{c}{MP3D (val)} \\ \cmidrule{3-5}
 &  & \multicolumn{1}{l}{Success \textuparrow} & \multicolumn{1}{l}{SPL \textuparrow} & \multicolumn{1}{l}{DTS \textdownarrow} \\ \midrule
\textbf{\texttt{BC}} & \redx & 3.8 & 2.1 & 7.5 \\\midrule
\textbf{\texttt{DDPPO}}\cite{DDPPO} & \redx & 8.0 & 1.8 & 6.9 \\
\textbf{\texttt{Red-Rabbit}}\cite{RedRabbit} & \redx & 34.6 & 7.9 & - \\
\textbf{\texttt{THDA}}\cite{THDA} & \redx & 28.4 & 11.0 & 5.6 \\\midrule
\textbf{\texttt{FBE}}\cite{FBE} & \redx & 22.7 & 7.2 & 6.7 \\
\textbf{\texttt{ANS}}\cite{ANS} & \redx & 27.3 & 9.2 & 5.8 \\
\textbf{\texttt{PONI}}\cite{PONI} & \redx & 31.8 & {12.1} & 5.1 \\ \midrule
\textbf{\texttt{ANS + SI}}\cite{Online} & \redx & 27.9 & 13.1 & 6.1 \\
\textbf{\texttt{SemExp + SI}}\cite{Online} & \redx & 34.7 & \textbf{15.1} & 5.8 \\ \midrule
\rowcolor{LightCyan}\textbf{\texttt{SEA}} (ours) & \brightgreencheck & \textbf{39.0} & {13.7} & \textbf{5.0}\\
\textbf{\texttt{SEA w/o Update}} & \brightgreencheck &  {33.3} & {13.6} & {5.7}\\ 
% $\textbf{\texttt{SEA + GT}}^{\dag}$ & \brightgreencheck & {62.0} & {23.8} & {3.3}\\ 
%  {33.3} & {13.6} & {5.7}
\bottomrule
%  {62.0} & {23.8} & {3.3}
% \textbf{\texttt{SEA}} (ours) & \brightgreencheck & \textbf{36.4} & \textbf{14.2} & \textbf{5.3}\\ \bottomrule
% \textbf{\texttt{SEA}} (ours) & \brightgreencheck & \textbf{36.7} & \textbf{13.6} & \textbf{5.3}\\ \bottomrule
% \textbf{\texttt{SEA}} (ours, noise) & \brightgreencheck & \textbf{34.1} & {8.6} & \textbf{5.6}\\ 
\end{tabular}
\label{tab:objectgoal_mp3d}  
}
% \end{minipage}
\end{table}

\paragraph{SEA outperforms navigation baselines} 
{\color{black}
Our SEA method sets a new benchmark by surpassing all previous state-of-the-art baseline methods on the MP3D dataset's validation split, as detailed in Table~\ref{tab:objectgoal_mp3d}. This achievement encompasses end-to-end reinforcement learning (RL), metric map-based baselines, and graph map-based methods, particularly in terms of Success and DTS metrics. Remarkably, SEA demonstrates a staggering 926.3\% improvement in Success compared to the behavior cloning model. Furthermore, SEA outperforms the \texttt{PONI}\cite{PONI} by 22.64\% in Success and by 13.20\% in SPL, thanks to its development of more efficient pathways. 
While \texttt{PONI}\cite{PONI} is highly reactive and excels in exploration with its frontier-based path generation, it falls short in exploitation; in contrast, SEA uses topological maps for long-term planning, allowing it to create more effective routes through better exploitation.
% 
% Although \texttt{PONI}\cite{PONI} employs a frontier-based approach for path generation in object searching, making it highly reactive, it excels primarily as an exploration model and falls short in terms of exploitation. In contrast, SEA leverages topological maps as explicit memory for long-term planning, which enables it to develop more optimal routes through enhanced exploitation capabilities.
% 
When compared to the \texttt{SemExp + SI}\cite{Online} model, which combines an abstract model with a semantic exploration approach, SEA increases Success by 12.4\%. This is particularly notable given that SEA does not use global pose information and is trained without interactive reinforcement learning. 
SEA outperforms \texttt{SemExp}\cite{Online} in terms of Success but has slightly lower SPL (by 1.4 percentage points) due to its adaptability. This adaptability allows SEA to eventually locate the object, even if it initially follows inefficient routes. In contrast, \texttt{SemExp}\cite{Online} relies on a highly accurate metric map and a perfect pose sensor, making it very efficient at following the optimal path. However, if \texttt{SemExp}\cite{Online} is led astray onto an incorrect path, it lacks the mechanism to update and correct itself, making it difficult to locate the object. SEA’s ability to adapt and correct its course enables higher success rates, even though this sometimes means taking longer and less efficient paths, resulting in lower SPL performance.
% 
% These findings bolster the hypothesis that effectively solving sub-problems related to place and place relationships is key for successful object goal navigation.

% However, despite its higher success rates, SEA lags behind SemExp~\cite{Online} when evaluated using the SPL (Success weighted by Path Length) metric. SPL measures the efficiency of a trajectory by comparing the agent’s path to the optimal path length when the goal is successfully reached. In typical environments, topological planning often follows a sequence of areas like living room $\rightarrow$ kitchen $\rightarrow$ bedroom $\rightarrow$ bathroom, reflecting a standard home layout. But if the test environment deviates from this typical structure, the probability distribution and planning approach must be adjusted to fit the new layout. This adaptation could result in less efficient paths but ensures that the goal is eventually reached.

% In summary, while SEA achieves higher success rates by adapting to the environment—even if it means temporarily following inefficient routes, SemExp~\cite{Online} remains highly efficient on the correct path but fails to adapt if it gets led astray, resulting in a lower SPL. This trade-off explains why SEA achieves state-of-the-art results in terms of success rate but falls short compared to SemExp~\cite{Online} in terms of SPL.
}
% Our SEA method outperforms all previous state-of-the-art baseline methods on the validation split of the MP3D dataset, as shown in Table~\ref{tab:objectgoal_mp3d}. This includes end-to-end RL and metric map-based baselines, as well as graph map-based methods, in terms of Success and DTS metrics. Compared to the behavior cloning model, SEA shows a 926.3\% improvement in Success. Additionally, SEA outperforms the $\textbf{\texttt{PONI}}$\cite{PONI} method by 22.64\% in Success and 13.2\% in SPL. 
% % 
% To analyze this outcome, it is crucial to understand the fundamental differences between the two methods. $\textbf{\texttt{PONI}}$\cite{PONI} employs a frontier-based approach to generate paths for object searching, which can be considered highly reactive. While $\textbf{\texttt{PONI}}$\cite{PONI} serves as an effective exploration model, it does not perform as well in terms of exploitation. In contrast, the proposed method leverages topological maps as an explicit memory for long-term planning, enabling it to devise more efficient paths through its exploitation capabilities.

% Compared to the model with the abstract model attached to the semantic exploration method, $\textbf{\texttt{SemExp + SI}}$\cite{Online}, Success is increased by 12.4\%. These results are surprising considering that global pose information is not used and the method is learned without interactive reinforcement learning, supporting the hypothesis that solving sub-problems related to place and place relationships is crucial for object goal navigation. 

\begin{figure}[!t]
  \centering
  \begin{minipage}[b]{0.45\linewidth}
    \centering
    \includegraphics[width=\linewidth]{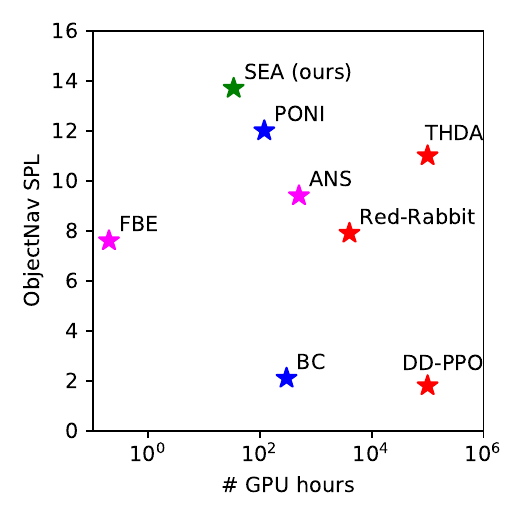}
    \caption{{Training cost.}}
    \label{fig:training_cost}
  \end{minipage}
  \hfill
  \begin{minipage}[b]{0.5\linewidth}
    \centering
    \includegraphics[width=\linewidth]{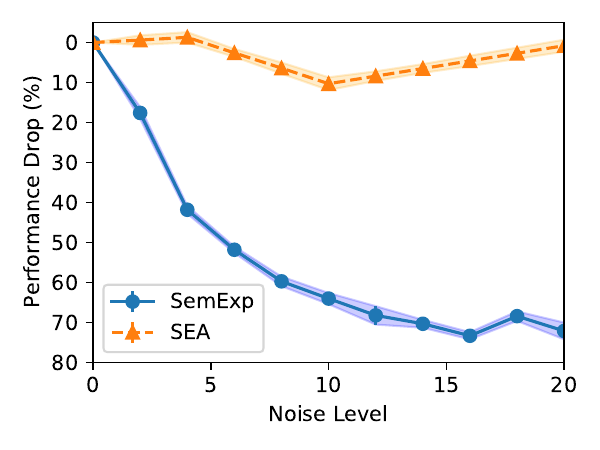}
    \centering\caption{{Ablation study on pose sensor noises.}}
    \label{fig:noise}
  \end{minipage}
\end{figure}

\paragraph{SEA has low computational requirements}
The place encoder can be trained within a day using a single GPU. Constructing SEA takes about 10 hours with a single GPU. During the inference, a GPU with 3000 MB memory is enough to run the trained encoder and detector. Our SEA has the lowest cost, three times less than the non-interactive SoTA baseline~\cite{PONI}, as demonstrated in Figure~\ref{fig:training_cost}.

\paragraph{SEA is robust to pose noises}

SEA demonstrates enhanced robustness to pose sensor noise by utilizing place reachability for long-term planning instead of building a metric map. As shown in Figure~\ref{fig:noise}, SEA experiences a modest performance drop in Success, with a 10\% decrease at a noise level of 10 and only a 0.77\% reduction at noise level 20, despite significant pose sensor interference. Here, the noise levels are indicative of real-world scenarios, with noise level 1 mimicking common robotic system disturbances and higher levels representing more severe interference. In contrast, the SemExp model shows a marked decline in efficiency—42\% at noise level 4 and 64\% at noise level 10, further deteriorating with higher noise levels. This emphasizes SEA's ability to maintain efficiency through consistent replanning, leveraging its global policy model effectively, even when local paths are obscured by noise. These results highlight SEA's superior adaptability over traditional metric map-based approaches that rely on precise pose sensors.
% SEA's robustness to pose sensor noise is achieved through its use of place reachability for long-term planning, as opposed to building a metric map. This is demonstrated in the performance drop on Success, as shown in Figure~\ref{fig:noise}, when compared to the result with no noise on the pose sensor (noise level 0). The noise level of 1 is representative of the noise present in real-world robotic systems. Since the baseline SemExp~\cite{SemExp} model is only pretrained on the Gibson~\cite{GibsonEnv} dataset, the results for SemExp~\cite{SemExp} are based on the Gibson dataset.
% When encountering mild pose sensor interference (noise level 4), our model's efficacy remains stable, while SemExp's efficiency declines by 42\%. With more pronounced pose sensor interference (noise level 10), SEA's efficiency decreases by 10\%, in stark contrast to SemExp's sharp 64\% decline.
% % 
% As the pose sensor interference intensifies further (noise level 20), SemExp's efficiency continues to deteriorate markedly, while SEA's outcome slightly reduces by only 0.77\% (from 39.0\% to 38.7\% in terms of success). This can be attributed to the global policy model's capacity to re-engage in global planning when the local policy cannot ascertain a path due to noise, thereby boosting overall efficiency through consistent replanning.
% % 
% The findings underscore SEA's enhanced resilience to pose sensor interference, compared to metric map-based approaches, which rely heavily on precise pose sensors.

\begin{table}[!t]
\caption{\small{Localization results.}}
\resizebox{\linewidth}{!}{
\begin{tabular}{cc|ccc} \toprule 
Method & Input & \multicolumn{1}{l}{Acc@0.5m \textuparrow} & \multicolumn{1}{l}{Acc@1m \textuparrow} \\ \midrule
\textbf{\texttt{NRNS}}~\cite{NRNS} & I  & 10.5 & 66.5 \\ 
\textbf{\texttt{VGM}}~\cite{VGM} & I  & 36.9 & 62.7 \\ 
\textbf{\texttt{TSGM}}~\cite{TSGM}  & I+O  & 38.9 & 65.1\\
\textbf{\texttt{SEA}} & I+O+P & \textbf{40.4} & \textbf{73.1} \\ \bottomrule
% \begin{tabular}{cc|ccc} \toprule 
% Method & Input & \multicolumn{1}{l}{Acc @ 0.5m \textuparrow} & \multicolumn{1}{l}{Acc @ 1m \textuparrow} & \multicolumn{1}{l}{Error \textdownarrow} \\ \midrule
% \textbf{\texttt{NRNS}}~\cite{NRNS} & I  & 29.2 & 11.9 & 6.1 \\ 
% \textbf{\texttt{VGM}}~\cite{VGM} & I  & 36.9 & 62.7 & 0.93 \\ 
% \textbf{\texttt{TSGM}}~\cite{TSGM}  & I+O  & 39.7 & 65.5 & 0.86 \\
% \textbf{\texttt{SEA}} & I+O+P & 36.4 & 73.1 & 0.96 \\ \bottomrule
\end{tabular}
}
\label{tab:localization}  
\end{table}

\paragraph{Semantic information helps to improve localization} 
{\color{black}
The accumulated graph data serves as a form of memory that integrates various pieces of semantic information. To determine whether this accumulated semantic information is indeed beneficial, we employ a trained network with attached probes to evaluate its utility for localization purposes. The localization probe network comprises Graph Neural Networks (GNNs) and transformer decoder networks. These networks are trained with the ground truth location coordinates (x, y). After training, the performance is assessed on a test set by calculating the distance between the predicted and actual locations. 
% If the calculated distance falls within an accuracy range (e.g., 0.5m or 1m), an accuracy score of 1 is assigned; otherwise, a score of 0 is given.
}

Upon analyzing the localization results {\color{black}(Table~\ref{tab:localization},} it is evident that our SEA method, utilizing inputs images (I), objects (O), and places (P), exhibits superior performance in both Acc@0.5m and Acc@1m metrics with scores of 40.4 and 73.1, respectively. The Acc@1m metric signifies the average accuracy of distance calculations within a 1-meter range. If the calculated distance is accurate within this range, an accuracy score of 1 is assigned, while inaccuracies are denoted as 0.
When the performance enhancement of SEA is calculated in terms of percentage increase, we observe substantial improvements over other methods. Specifically, compared to NRNS~\cite{NRNS}, SEA demonstrates an extraordinary increase of approximately 285\% in Acc@0.5m. In comparison to VGM~\cite{VGM}, SEA shows a noticeable improvement as well. The TSGM~\cite{TSGM} method, with inputs I and O, ranks second to SEA, yet SEA still surpasses it in terms of accuracy.
To summarize, these results not only indicate the dominance of SEA in localization accuracy, particularly within a 1-meter range, but also emphasize the considerable performance enhancement achieved by incorporating semantic knowledge.

% \hfill % adds horizontal space between minipages

% \begin{minipage}{0.45\textwidth}

\begin{table}[!t]
% \vspace{10pt} % adds vertical space between the tables
\centering
\caption{{Impact of place info.}}
% Place-based subgoal / Place-based stop  Ablation study on the i
\resizebox{\linewidth}{!}{
\begin{tabular}{cc|ccc} \toprule 
\multicolumn{2}{c}{\textbf{\texttt{SEA}} ablations} & \multicolumn{3}{|c}{MP3D (val)} \\ \midrule
Subgoal & Stop & \multicolumn{1}{l}{Success \textuparrow} & \multicolumn{1}{l}{SPL \textuparrow} & \multicolumn{1}{l}{DTS \textdownarrow} \\ \midrule
\redx  & \redx  & 29.2 & 11.9 & 6.1 \\ %sixtail
\brightgreencheck  & \redx  & {36.4} & {14.2} & {5.3}\\
\brightgreencheck  & \brightgreencheck & {39.0} & {13.7} & {5.0}\\ \bottomrule % lapras
\end{tabular}
}
\label{tab:ablation_study_on_place}  
% \end{minipage}
% 
\end{table}

\paragraph{Relation update is effective on adapting to unseen environment}
The effectiveness of adapting to the unseen environment through relation updates is demonstrated in the results of an ablation study presented in the second row from the bottom of Table~\ref{tab:objectgoal_mp3d}. As new place connections or place-object connections arise during testing, the probability distribution is updated at each step of the current episode. This acquired connection information is utilized solely within the episode and is not stored for future use.
The experiment yielded a 17.1\% increase in success rate compared to the case where no episodic update was applied. Notably, SPL only improved by 0.7\%. This is likely due to the episodic update altering the posterior distribution. The agent initially navigates to the location where the target object is most likely to be found and checks for its presence. If the object is not discovered in the initial place cluster, the agent may become stuck. However, the relation update weakens the connection between the target place cluster and the target object, allowing the agent to replan and reach the next most connected place cluster. As the agent successfully finds the object by moving to the next place cluster, SPL decreases as the path length of the successful path becomes longer.

\paragraph{Place cluster is useful for planning}
We evaluated the impact of place-specific information in the planning process using a model that randomly designates subgoals. In this context, "Subgoal" refers to the place-based subgoal selection method, which strategically enhances reachability based on place connections.
As delineated in Table~\ref{tab:ablation_study_on_place}, the implementation of this subgoal selection method resulted in quantifiable improvements. Specifically, success rates increased by 24.7\%, and the SPL metric concurrently rose by 19.3\%.
The term "Stop" denotes the place-based stop mechanism, a model variant that facilitates termination before reaching the predetermined target place cluster. The examination of this mechanism revealed distinct effects. The success rate decreased by 7.1\%. In conclusion, the use of place clustering proves to be advantageous for the planning process.

\begin{figure*}[t!]{\centering\includegraphics[width=\textwidth]{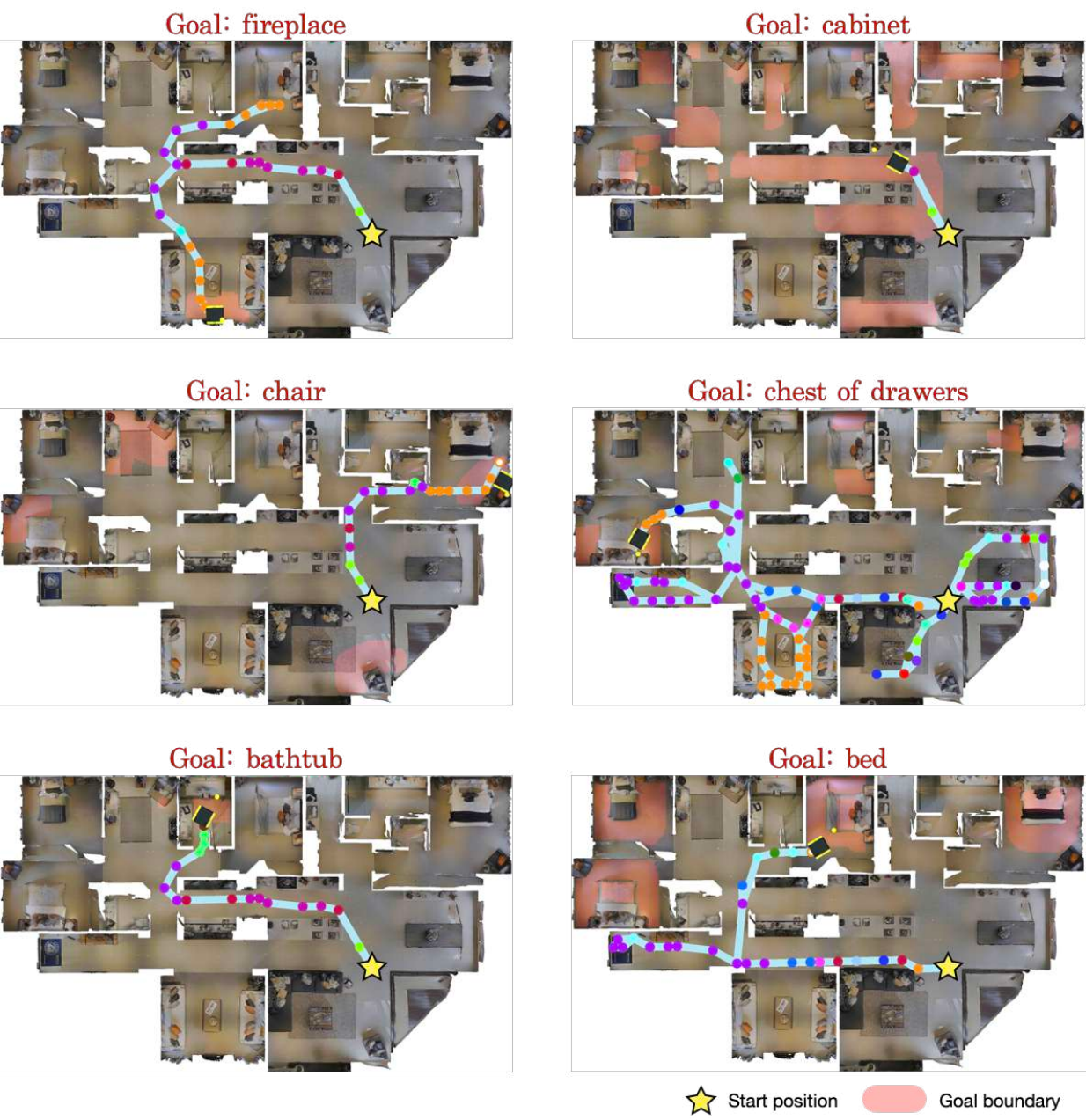}}\centering
\caption{
\textbf{Learning semantic relationships.}
% The figure provides an example demonstrating how SEA can identify efficient paths by leveraging semantic relationships. 
% % It depicts trajectories for six object goals: fireplace, cabinet, chair, chest of drawers, bathtub, and bed. 
% The trajectories of these six target object categories illustrate various pathways that are adapted according to the specific target object. 
% The starting point is symbolized by a yellow star, while a red region represents the goal boundary, which denotes that the goal is within a 1m radius.
The figure illustrates how SEA can identify efficient paths by leveraging semantic relationships. It shows the trajectories for six target objects: a fireplace, a cabinet, a chair, a chest of drawers, a bathtub, and a bed. These trajectories demonstrate various pathways that are adapted to each specific target object. The starting point is marked by a yellow star, while the goal boundary is represented by a red region, indicating that the goal is within a 1-meter radius.
% The starting point is marked with a yellow star, while the goal boundary, indicating that the goal is within a 1m range, is represented by a red region.
}
\label{fig:goal_ablation}
\end{figure*}

\begin{figure*}[t!]{\centering\includegraphics[height=0.85\textheight]{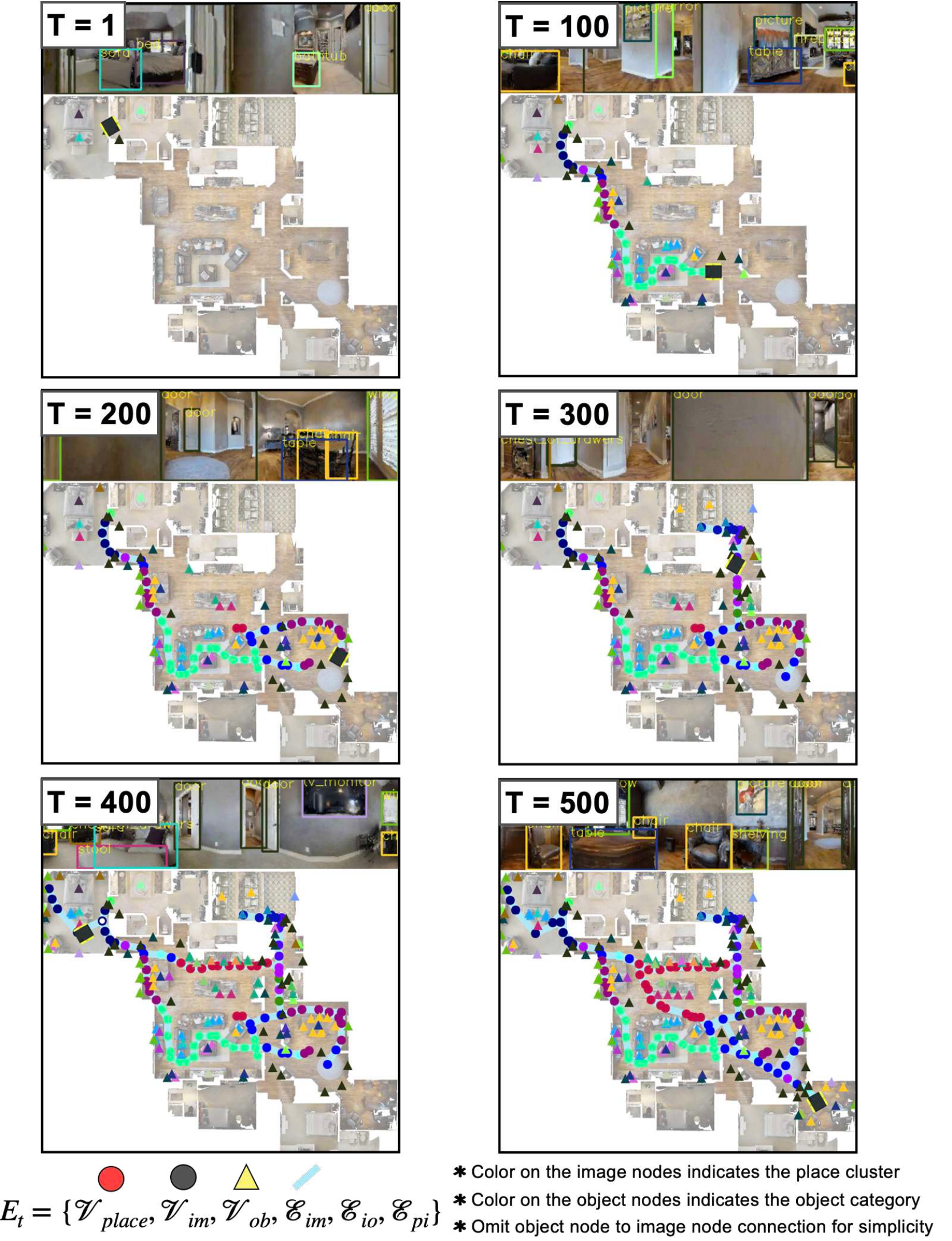}}\centering
\caption{
\textbf{Example of constructing an episodic graph.}
While randomly investigating the surroundings, the agent accumulates an episodic graph. A place graph, an image graph, and an object graph comprise an episodic graph.
Color on the image node indicates the place cluster, circles indicate image nodes, and triangles represent object nodes.
% % 
% The reachability and connectedness between places and objects are collected when exploring. 
% % 
% If the place clusters of two connected image nodes changes, the two place clusters are connected.
% % 
% For example, when t=100, the agent moves from the bedroom (blue) to the kitchen (purple).
% % 
% The reachability between the blue and purple place clusters is one. 
% % 
% The topological semantic graph can obtain connectedness between nodes Without utilizing pose information.
% % 
% Top-down maps and node positions are only used for visualization and are not used to construct a graph.
}
\label{fig:episodic_graph_example}
\end{figure*}

\begin{figure*}[t!]{\centering\includegraphics[width=0.8\textwidth]{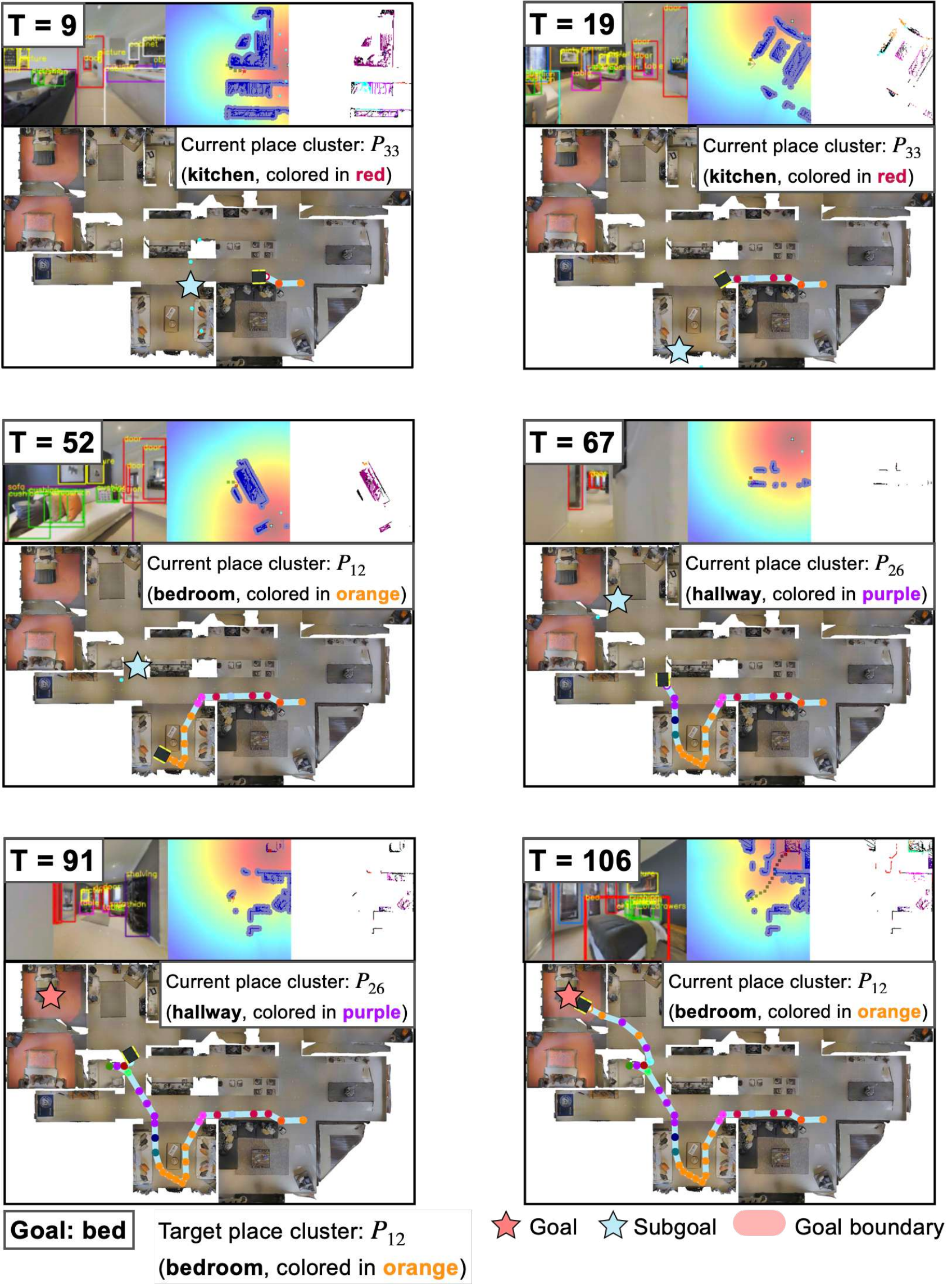}}\centering
\caption{
\textbf{Qualitative examples of navigation using SEA.}
When given the goal object as a bed, the agent formulates a plan to navigate to the bedroom ($P_{12}$). To reach the bedroom ($P_{12}$), the agent predicts the subgoal place cluster using the categories of the initially visible objects.
% 
% \textbf{Row 1:} Initially, the agent estimates that the bedroom ($P_{12}$) is to the left, with the kitchen positioned directly ahead and to the right. It selects the left subgoal, anticipating that the target place is in that direction.
% % 
% \textbf{Row 2:} Upon reaching the subgoal, the agent commences a search for a bed, but fails to locate one. The agent then exits the area to identify a new subgoal. Given the visible door on the left, it deduces that the target place likely exists in that direction.
% % 
% \textbf{Row 3:} As the agent progresses towards the new subgoal, it detects the target object (a bed). The agent then devises a local plan to approach the target object. Finally, when the agent recognizes the matching place cluster as the searched-for bedroom ($P_{12}$), it activates the stop command.
% 
% We visualized more examples in the supplemental video.
} %\squeezeup
\label{fig:eval_example1}
\end{figure*}

{\color{black}
\paragraph{SEA can choose different trajectories based on different goals}
This experiment demonstrates that the path is contingent upon the chosen goal, with the starting point fixed and goal categories varied. In Figure~\ref{fig:goal_ablation}, the yellow star marks the starting point, and the red region indicates the goal boundary (1m radius), solely for visualization. 
The figure shows trajectories for six object goals: fireplace, cabinet, chair, chest of drawers, bathtub, and bed. Each trajectory adapts to its respective target. Detailed analysis is available in the supplementary material.

\section{Visualization}

\paragraph{Construction of SGMs}
The process of constructing a SGM is depicted in Figure~\ref{fig:episodic_graph_example}.
A SGM integrates place graphs, image graphs, and object graphs.
In this representation, image nodes are depicted as circles. Their colors correspond to the respective place clusters. Object nodes are depicted as triangles. The colors of these nodes indicate object categories, matching the colors of the bounding boxes in the panoramic RGB image.
For clarity, connections between image nodes and object nodes have been omitted in the visualization. It is important to note that top-down maps and the positions of image and object nodes are utilized solely for visualization purposes and are not used as input data.
Additionally, a supplementary video is available that demonstrates the construction of semantic graph maps, showcasing one episode across 20 training scenes.

\paragraph{Example visualization of episodes}
We provide an example visualization of an episode in Figure~\ref{fig:eval_example1}. This shows how semantic prior graphs are used in the global policy to perform ObjectNav, where the goal is identified as 'bed'.
The navigation process begins with the agent perceiving the bedroom ($P_{12}$) to be on the left side and the kitchen to be at the front and right side. Based on this initial perception, the agent decides to move left, anticipating that the target location might be there. Upon reaching the subgoal, the agent searches for a bed but does not find one. Consequently, the agent exits the current location and re-evaluates the subgoal, noticing a door on the left and inferring that the target location is likely on the left side. While proceeding towards the subgoal, the agent eventually encounters the target object, the bed. The agent then formulates a local plan to reach the bed. Finally, after confirming that the location matches the intended destination, the bedroom ($P_{12}$), the agent presses the stop button. Detailed information and additional examples are provided in the supplementary material.

}

\section{Conclusions and Future Work}
\subsection{Conclusions}
We present SEA, a method for learning semantic relationships between places and objects in unknown environments with low computational cost. Our approach identifies the object's location and navigates using place connections. By adapting to the unseen environment through relation updates, SEA achieves state-of-the-art results for ObjectNav in MP3D. Unlike other methods {\color{black}are} not robust to noise pose sensor, SEA is robustly navigate environment with noisy settings. Our method is the first to not require a global metric map for ObjectNav in large environments like MP3D. We show that incorporating semantic relationships improves localization and navigation tasks. 
\subsection{Future work}
{\color{black} In addition to the current approach, we propose future directions and still open challenges: 

1. \textbf{Incorporating language features for 3D objects}: Our proposed method relies solely on image features to define objects. Using language features for 3D objects could lead to more general features that can improve object search and correlation graph connections in the metric space. 

2. \textbf{Agents in dynamic environments}: Recognizing changes in the environment, such as a cup being moved from the kitchen to the living room, can aid in task-solving. If an agent maintains memory in the form of a graph, it can adapt much better to dynamic environments compared to using a metric map. 

3. \textbf{Recognizing physical laws in the environment}: To manipulate objects or perform meaningful control tasks, it is necessary to understand the physical laws governing the environment. For example, avoiding small wooden blocks on the floor or considering the center of mass when picking up a tool. 

4. \textbf{Interactive intelligence}: Our method should not only rely on its own intelligence but also interact with humans or other robots in the environment to update the topological map. Further advancements in these areas can lead to more robust and effective navigation and localization in complex environments. 

We believe that our proposed method is a step towards achieving this goal, and we look forward to future developments and improvements. Additional details can be found in the supplementary material.
}

\clearpage

\clearpage

\clearpage

\clearpage
% \chapter{Supplementary Material}
% \chapter{Supplementary Material}

\appendix

This document provides additional information about experimental settings and includes supporting qualitative visualizations.
We recommend watching our supplementary three-minute video, which showcases animated examples of navigation using SEA.
Here is an overview of the sections included in the supplementary file:
% Below is a summary of the sections in the supplementary file:
\begin{itemize}
  \item \ref{supp:Implementation}: Additional Experimental Details
  \item \ref{supp:Episode}: Examples of Semantic Graph Maps %/ Visualization of prior distributions  
  \item \ref{supp:placegraph}: Visualization of Place Clusters and Reachability
%   \item (§S\ref{supp:disbributions}) 
  \item \ref{supp:Test}: Visualization of ObjectNav Episodes
\end{itemize}



\title{Semantic Environment Atlas for Object-Goal Navigation}

% \affiliation[label1]{organization={},

% \affiliation[label2]{organization={},

\author{Nuri Kim, Jeongho Park, Mineui Hong, Songhwai Oh} %% Author name

\section{Additional Experimental Details}~\label{supp:Implementation}
\paragraph{Implementation details}
To supplement the main paper, we provide further details about the experiments. The MP3D~\cite{Matterport3D} encompasses 40 object categories, while the ObjectNav dataset from the Habitat challenge contains 21 specific object categories as goals. These categories are as follows:
\textit{chair}, \textit{table}, \textit{picture}, \textit{cabinet}, \textit{cushion}, \textit{sofa}, \textit{bed}, \textit{chest of drawers}, \textit{plant}, \textit{sink}, \textit{toilet}, \textit{stool}, \textit{towel}, \textit{tv monitor}, \textit{shower}, \textit{bathtub}, \textit{counter}, \textit{fireplace}, \textit{gym equipment}, \textit{seating}, and \textit{clothes}.
Considering some object categories in MP3D are not applicable for place-object statistics, we decided to utilize 32 specific object categories, as illustrated in Figure~\ref{fig:object_importance}. The following eight categories were excluded: \textit{wall}, \textit{floor}, \textit{ceiling}, \textit{column}, \textit{lighting}, \textit{beam}, \textit{railing}, \textit{misc}.
% Since some object categories in MP3D are not useful for place-object statistics, we opted to use 32 specific object categories, as detailed in Figure~\ref{fig:object_importance}. We did not use the following eight categories: \textit{wall}, \textit{floor}, \textit{ceiling},    \textit{column}, \textit{lighting}, \textit{beam}, \textit{railing}, \textit{misc}.
% 
MP3D comprises 30 ground truth places. However, due to the ambiguity of places such as \textit{other room} and \textit{junk}, we opted for eight evident places for place clustering: \textit{living room}, \textit{bedroom}, \textit{kitchen}, \textit{closet}, \textit{dining room}, \textit{bathroom}, \textit{toilet}, \textit{hallway}.
% There are 30 ground truth places in MP3D. Since some places, such as \textit{other room} and \textit{junk}, are ambiguous, we choose to use eight obvious places for place clustering: \textit{living room}, \textit{bedroom}, \textit{kitchen}, \textit{closet}, \textit{dining room}, \textit{bathroom}, \textit{toilet}, \textit{hallway}.

\begin{figure*}[t!]{\centering\includegraphics[width=\linewidth]{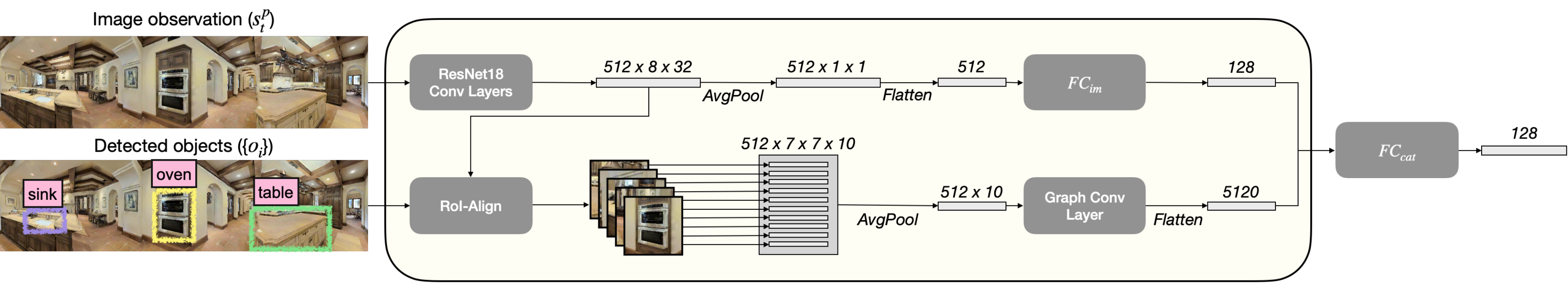}}\centering
\caption{
\textbf{Place encoder.} Figure showing the architecture of the place encoder.
}%\squeezeup\squeezelittle
\label{fig:place_encoder}
\end{figure*}

\begin{figure*}[t!]{\centering\includegraphics[width=\linewidth]{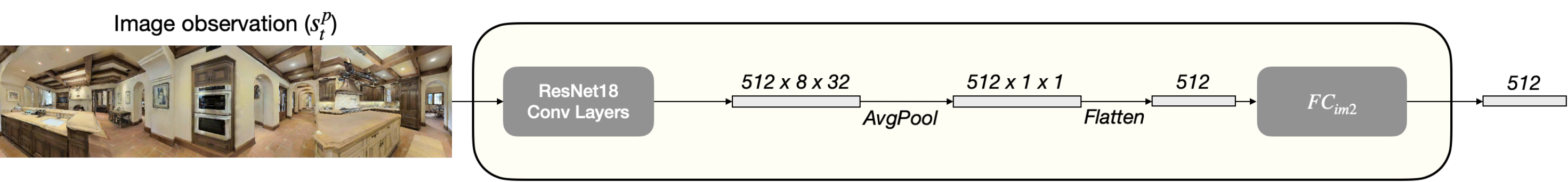}}\centering
\caption{
\textbf{Image encoder.} Figure showing the architecture of the image encoder.
}%\squeezeup\squeezelittle
\label{fig:image_encoder}
\end{figure*}

\begin{figure*}[t!]{\centering\includegraphics[width=\linewidth]{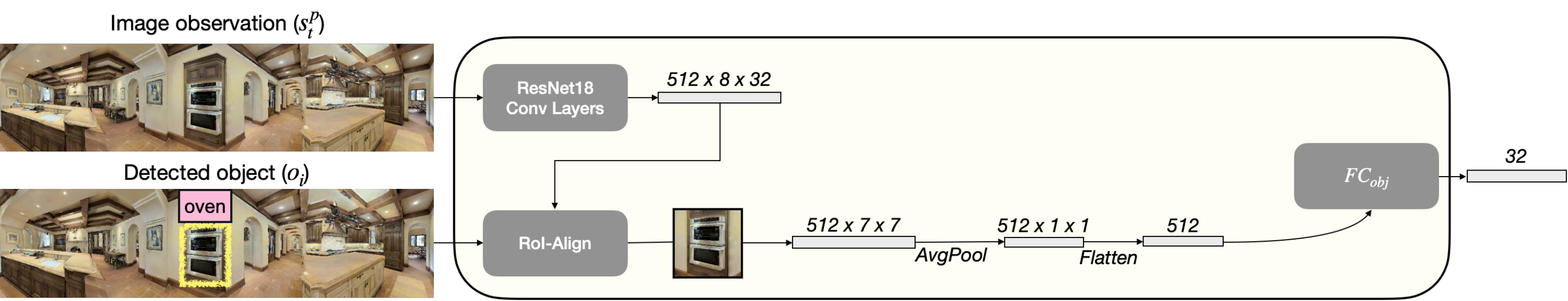}}\centering
\caption{
\textbf{Object encoder.} Figure showing the architecture of the object encoder.
}%\squeezeup\squeezelittle
\label{fig:object_encoder}
\end{figure*}

\begin{figure*}[ht]{\centering\includegraphics[width=\textwidth]{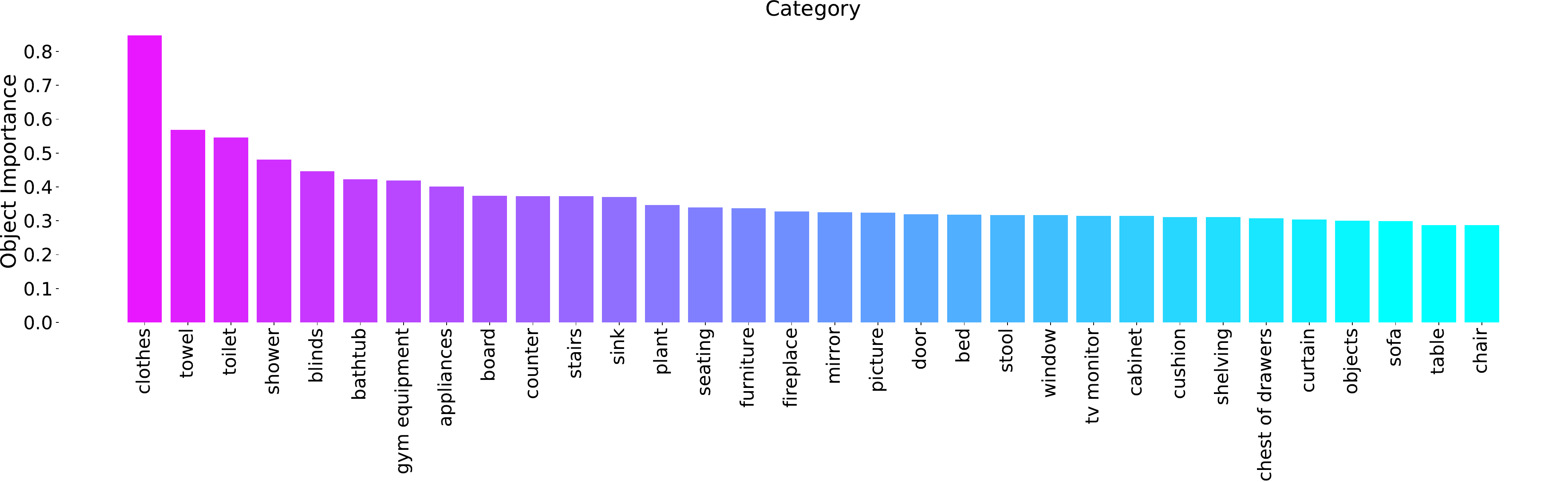}}\centering
\caption{Illustrating the significance of objects in distinguishing and identifying individual place clusters.}
\label{fig:object_importance}
\end{figure*}

\paragraph{Network parameters}
We employed deep learning exclusively to train three encoders: the place encoder, image encoder, and object encoder.
As depicted in Figures \ref{fig:place_encoder}, \ref{fig:image_encoder}, and \ref{fig:object_encoder}, the encoder networks utilize ResNet18~\cite{ResNet} as a backbone network and include fully connected layers.
For the place encoder in Figure~\ref{fig:place_encoder}, we use a single graph convolution layer for processing context of objects.
For the place encoder, the maximum number of the objects are set to 10.
The object features for place encoder and object encoder were extracted from images using RoI-align~\cite{he2017mask}.
% 
% A stochastic gradient descent (SGD) optimizer is utilized for training three encoders with a learning rate of 0.03, and the total training epoch is set to 200.
We utilized a stochastic gradient descent (SGD) optimizer to train the three encoders with a learning rate set at 0.03, over a total of 200 training epochs.
We randomly collected 5,000 panoramic RGB images per training environment for training encoders, totaling 305,000 images.
The objects were collected using a ground truth detector.
We randomly selected 20,000 images as a batch for each epoch, a feasible number for $K$-means clustering, with the mini-batch size set at 128.
In training the place encoder, we specified the temperature parameter ($\zeta$) in the clustering loss as the likelihood of an instance belonging to the corresponding cluster, following the approach in~\cite{PCL}.
The temperature parameter ($\zeta$) used in the metric learning was set to 0.2 for the place loss and 0.7 for the near loss.
The number of negative samples ($r$) is set to 128.
The number of place clusters ($\text{N}_p$) was set to 50, which is higher than the actual number of place labels to cover all spaces.
Place features have 128 dimensions, and image feature dimension is 512  ($D_i$=512). The object features have 32 dimensions ($D_o$=32).
$N$ is the total number of train environments, 61.
For the training image encoder, which is trained in an unsupervised manner, we used near loss and cluster loss.
Other parameters in the image encoder are the same as the place encoder.
To make the object encoder robust to an erroneous object detector, the positive sample for training the object encoder is augmented by randomly cropping the image of objects. 
As mentioned in the main manuscript, the similarity thresholds for constructing image and object graphs are set to 0.8, similarly used in~\cite{VGM, TSGM}.

%\squeezelittle
%\squeezelittle
%\squeezelittle

\paragraph{Selecting a subgoal}
Based on the current maximum depth value ($\text{d}_m$) of the directional camera sensor (which has a maximum depth of 5m), the local policy generates a local map.
% The local policy generates a local map based on the current maximum depth value ($\text{d}_m$) of the directional camera sensor, which has a maximum depth of 5m.
% 
When the size of the selected local map is $\text{d}_m$ $\times$ $\text{d}_m$, subgoal candidates for a local map at $\text{d}_m$ away are selected.
If the subgoal is beyond the straight line distance, it is considered unreachable and is eliminated, just like NRNS~\cite{NRNS}.
% 
% Then, the candidate that is reachable and closest to the selected semantic subgoal is selected as a local goal.
Subsequently, the candidate that is both reachable and closest to the selected semantic subgoal is chosen as the local goal.
The local policy's maximum number of steps is $2 \times \text{d}_m/0.4$.
For example, if the local policy produced a 3$m \times$ 3$m$ local map, the maximum local step size is 15, whereas a 5m local map has a maximum step size of 25.
If the agent reaches the local goal or exceeds the maximum step, the agent does global planning to select a new subgoal.
Furthermore, if the place clusters to which subgoals belong are all the same, or if no detected objects are found, a subgoal is chosen randomly.
% 
% This process allows an agent to explore environments more, preventing it from being stuck in a region.
This process facilitates greater exploration by an agent, thereby preventing it from getting stuck in a particular region.

\section{Illustrations of Semantic Graph Maps}~\label{supp:Episode}

\subsection{Prior distributions}%~\label{supp:disbributions} 
% The collected prior distributions from the semantic graph maps are visualized in the subsequent figures.
The following figures illustrate the prior distributions gathered from the semantic graph maps.

\begin{figure*}[t!]{\centering\includegraphics[width=\textwidth]{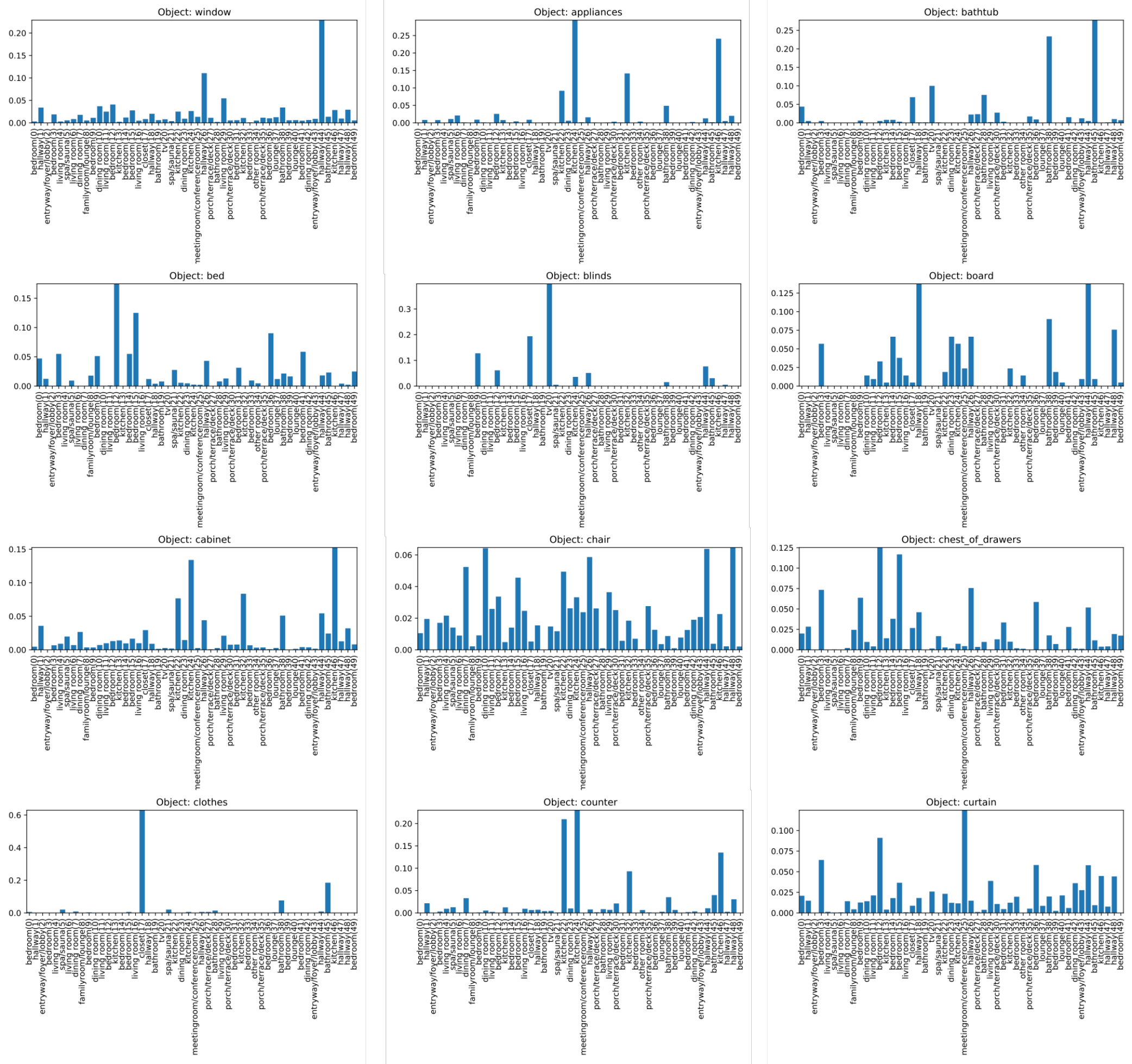}}\centering
\caption{{Place distribution given an object category ($p(\mathbb{P}|\mathbb{O})$).}}
\label{fig:place_distribution_1}
\end{figure*}

\begin{figure*}[t!]{\centering\includegraphics[width=\textwidth]{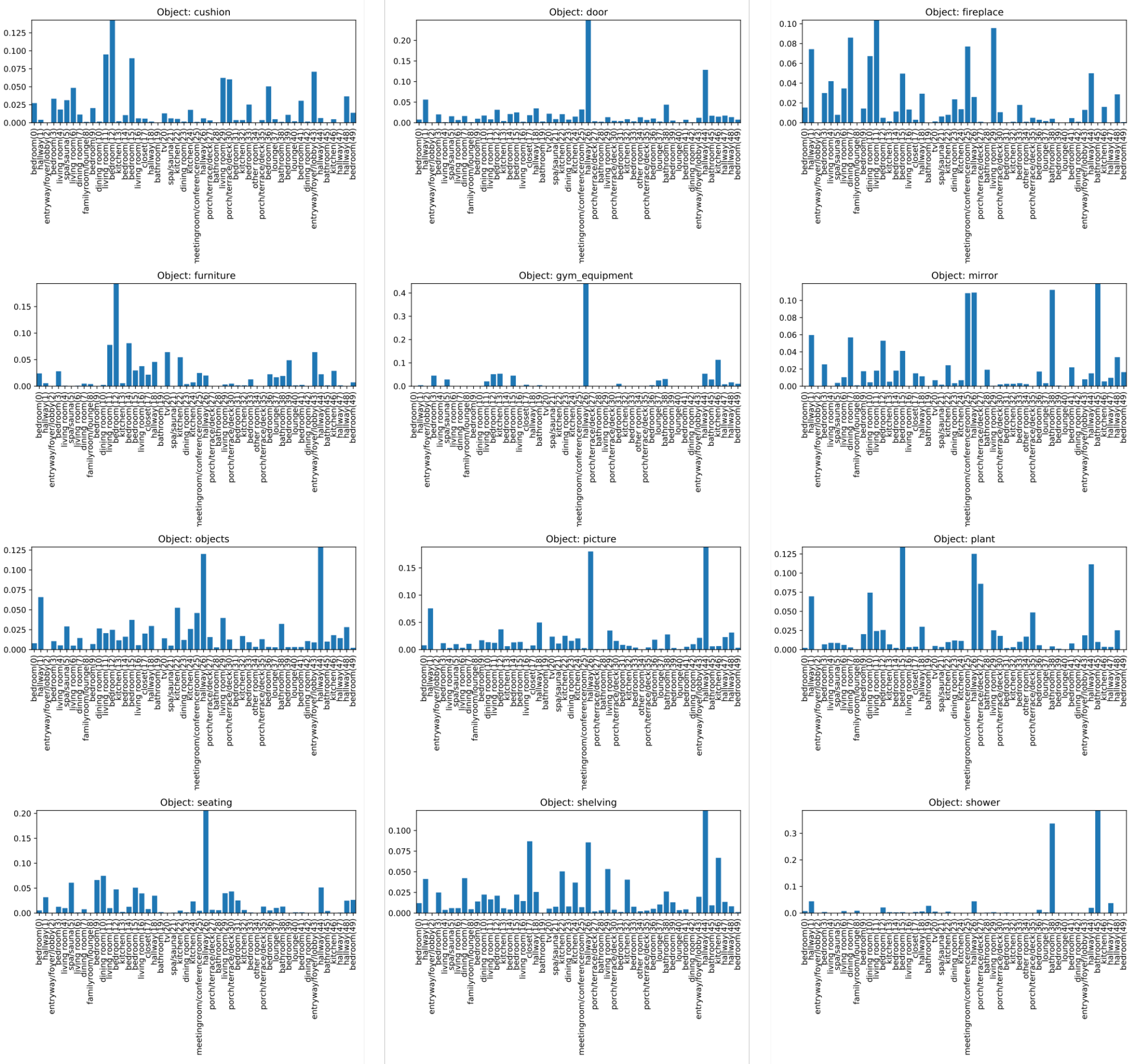}}\centering
\caption{{Place distribution given an object category ($p(\mathbb{P}|\mathbb{O})$).}}
% \caption{{Object distribution given place clusters ($p(\mathbb{O}|\mathbb{P})$).}}
\label{fig:place_distribution_2}
\end{figure*}

\begin{figure*}[t!]{\centering\includegraphics[width=\textwidth]{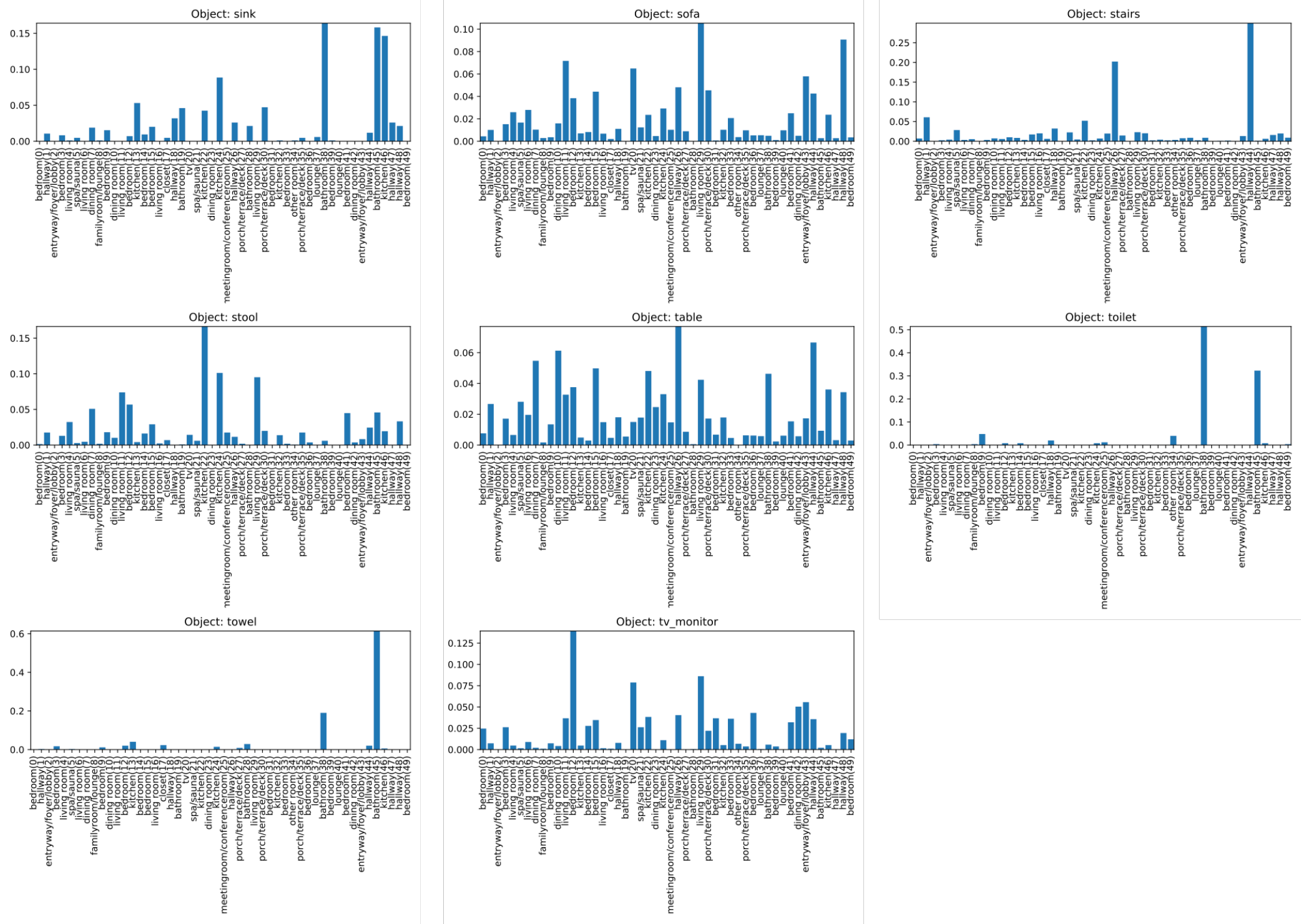}}\centering
\caption{{Place distribution given an object category ($p(\mathbb{P}|\mathbb{O})$).}}
% \caption{{Object distribution given place clusters ($p(\mathbb{O}|\mathbb{P})$).}}
\label{fig:place_distribution_3}
\end{figure*}

\paragraph{Place distribution}
$p(\mathbb{P}|\mathbb{O})$ is represented in Figures ~\ref{fig:place_distribution_1}, \ref{fig:place_distribution_2}, and \ref{fig:place_distribution_3}.
As depicted in Figure\ref{fig:place_distribution_1}, beds can be located in $P_{12}$ (bedroom), $P_{15}$ (bedroom), and $P_{36}$ (bedroom) respectively.
Clothes are primarily found in $P_{17}$ (closet), but can occasionally be found in $P_{45}$ (bathroom) due to the presence of shower robes.
Chairs are featured in a variety of spaces.
Figures\ref{fig:place_distribution_2} and ~\ref{fig:place_distribution_3} illustrate the strong correlation between places and objects.

\begin{figure*}[t!]{\centering\includegraphics[width=\textwidth]{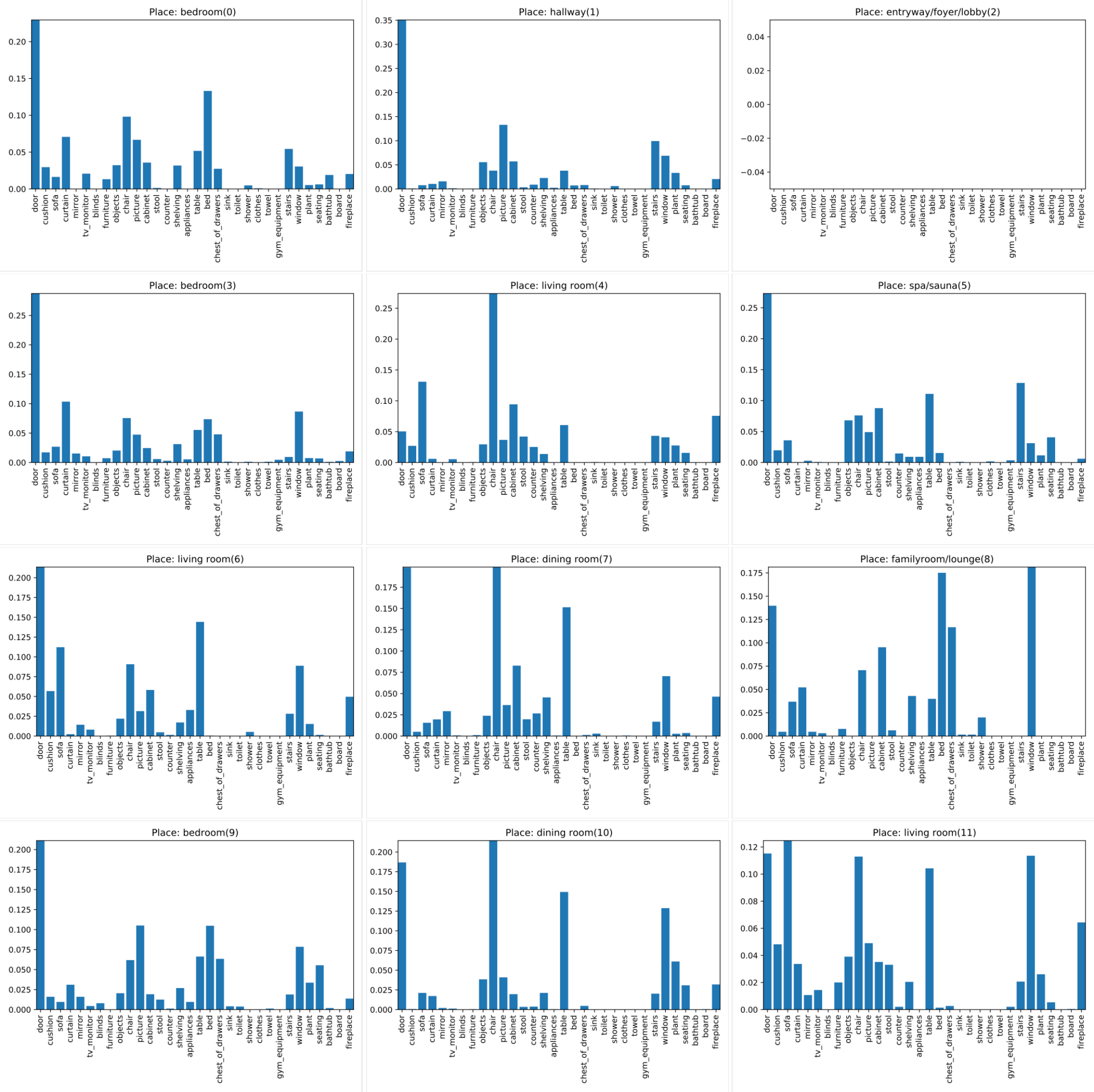}}\centering
\caption{{Object distribution given a place cluster ($p(\mathbb{O}|\mathbb{P})$).}}
\label{fig:object_distribution_1}
\end{figure*}

\begin{figure*}[t!]{\centering\includegraphics[width=\textwidth]{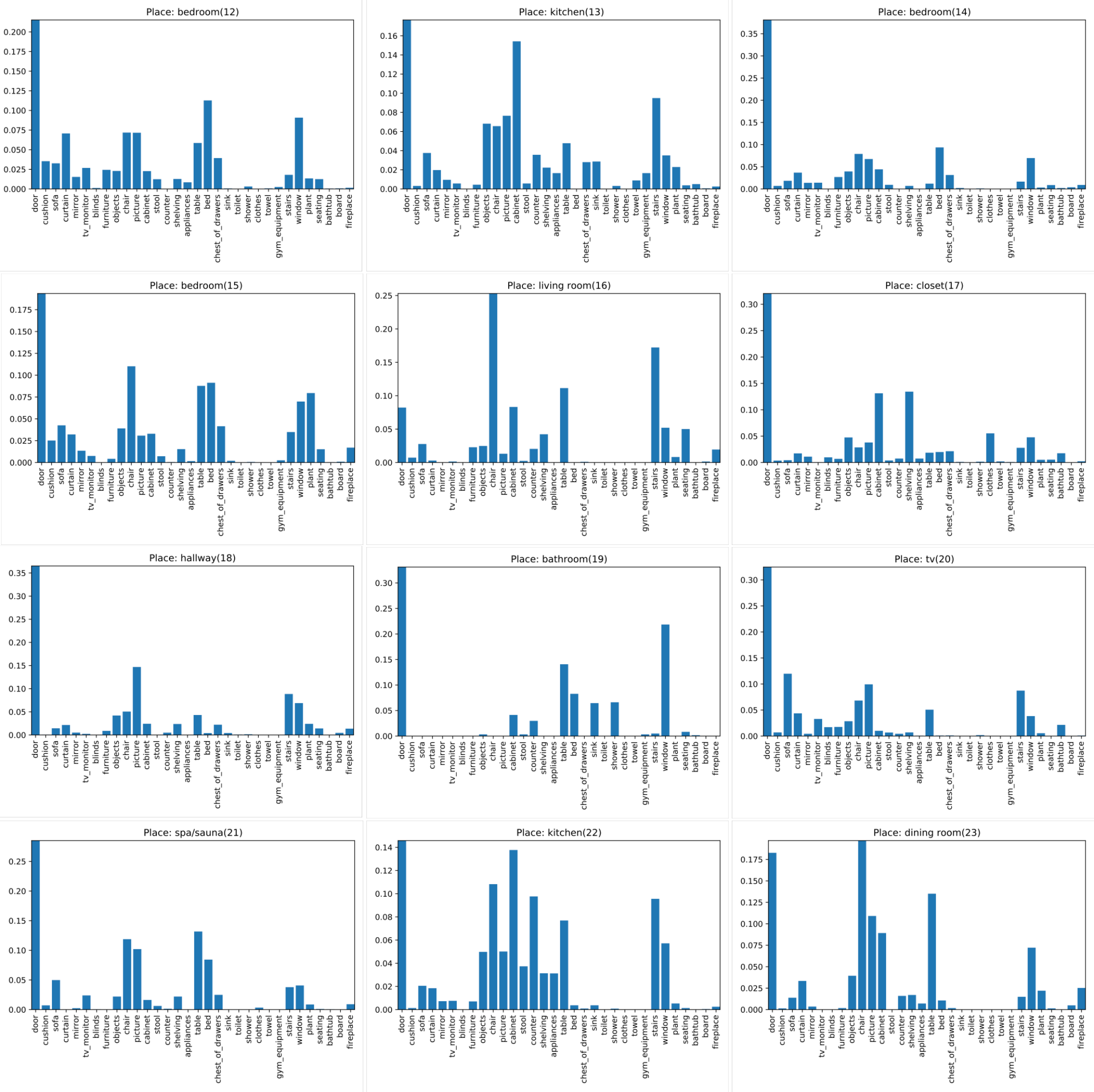}}\centering
\caption{{Object distribution given a place cluster ($p(\mathbb{O}|\mathbb{P})$).}}
\label{fig:object_distribution_2}
\end{figure*}

\begin{figure*}[t!]{\centering\includegraphics[width=\textwidth]{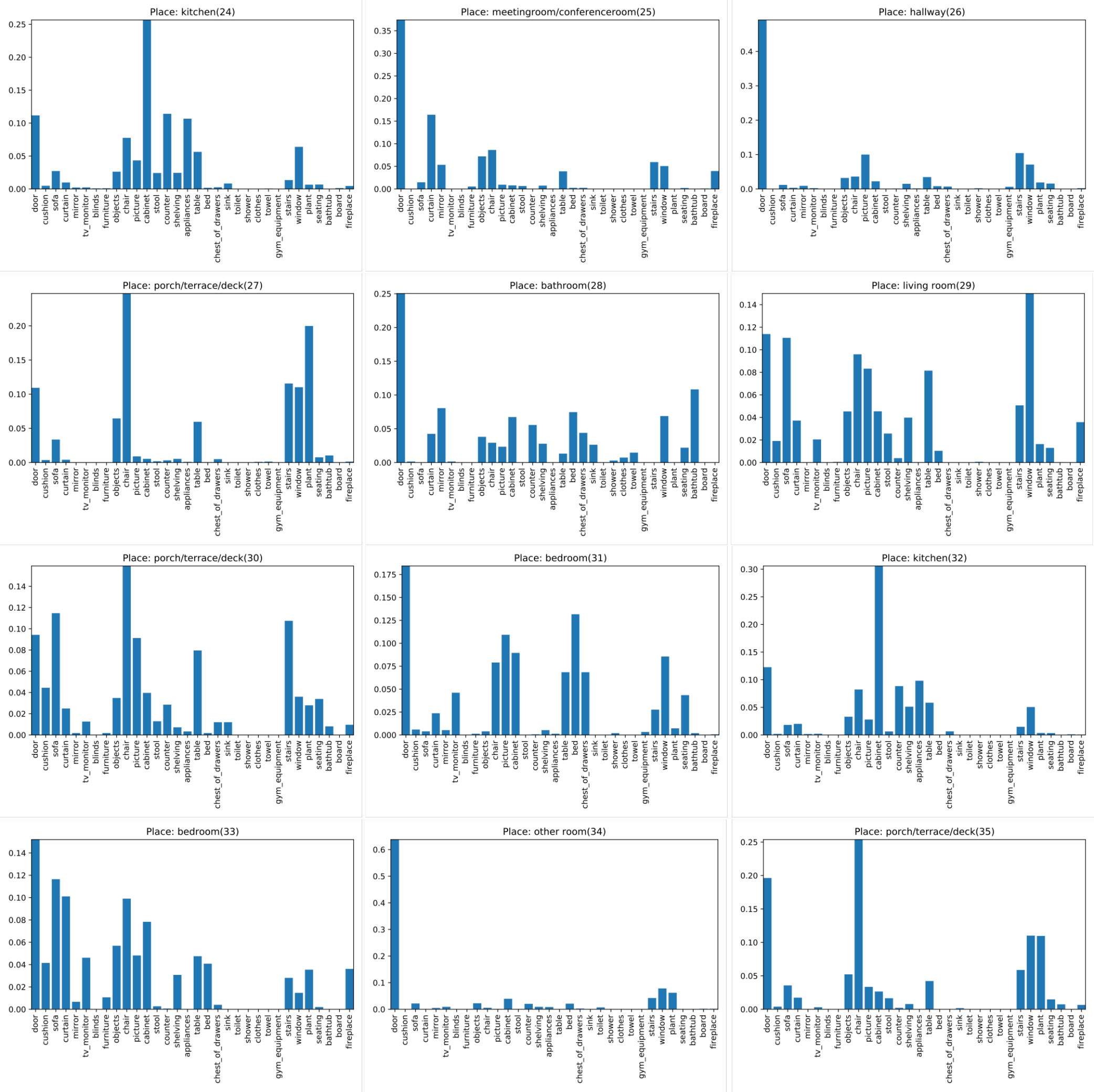}}\centering
\caption{{Object distribution given a place cluster ($p(\mathbb{O}|\mathbb{P})$).}}
\label{fig:object_distribution_3}
\end{figure*}

\begin{figure*}[t!]{\centering\includegraphics[width=\textwidth]{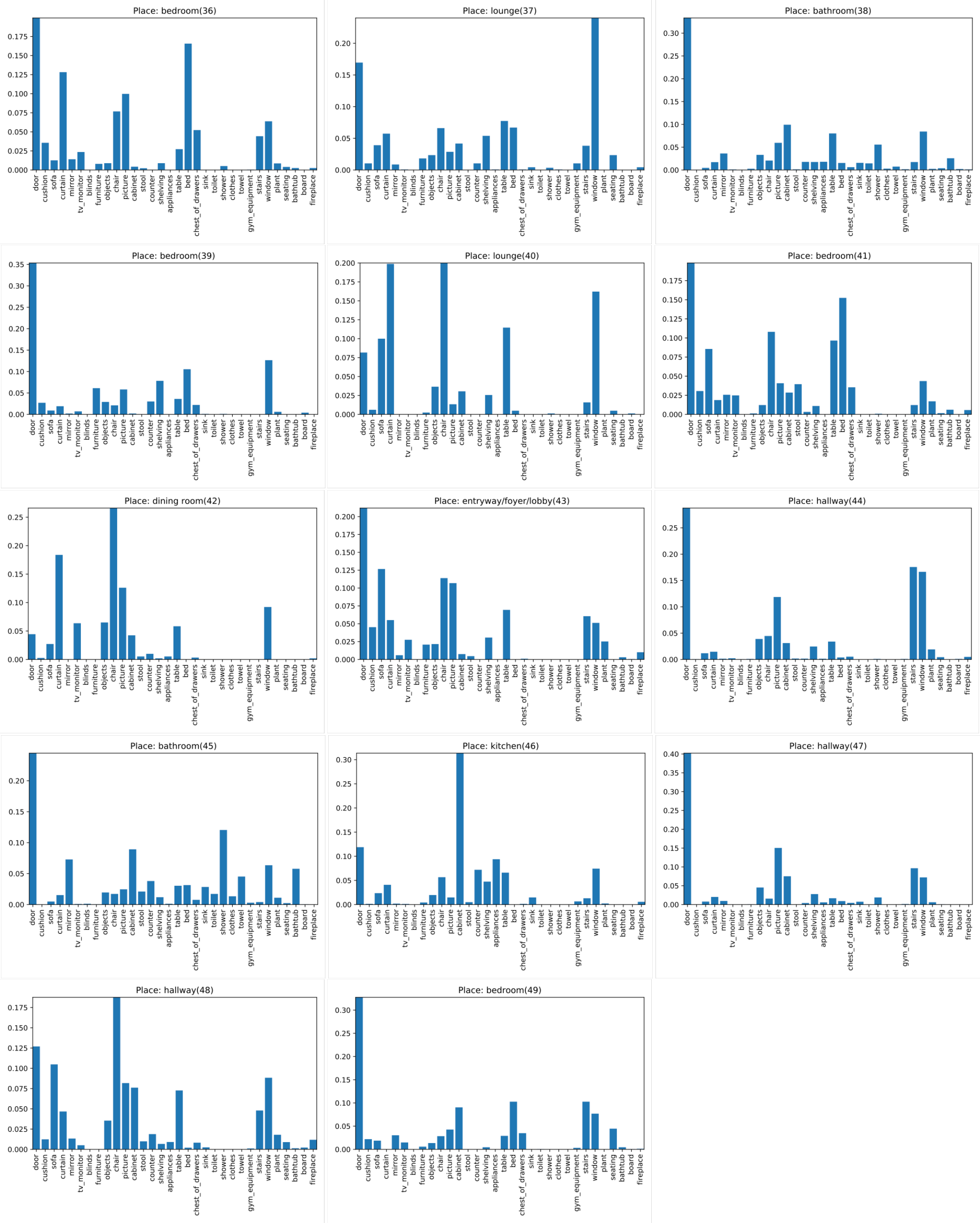}}\centering
\caption{{Object distribution given a place cluster ($p(\mathbb{O}|\mathbb{P})$).}}
\label{fig:object_distribution_4}
\end{figure*}

\paragraph{Object distribution}
$p(\mathbb{O}|\mathbb{P})$ is delineated in Figures~\ref{fig:object_distribution_1}, ~\ref{fig:object_distribution_2}, \ref{fig:object_distribution_3}, and \ref{fig:object_distribution_4}.
As shown in Figure\ref{fig:object_distribution_1}, $P_2$ is absent as the episode graphs couldn't entirely map the environment within ten episodes per scene.
A better coverage of the environment could potentially enhance SEA's performance.
%
% Doors appear quite frequently across all place clusters.
In Figure\ref{fig:object_distribution_4}, within $P_{41}$ (bedroom), the bed has the highest frequency, barring the door that is ubiquitous across all place clusters.
This demonstrates that the place clusters are trained to encapsulate various locations in accordance with object distribution.

\paragraph{Object importance}
Object categories are arranged in Figure~\ref{fig:object_importance} in order of their importance, defined as the inverse of the entropy of object distribution conditioned on places.
The importance values are derived from the prior distribution without any updates, and may change during navigation in a test environment.

\section{Visualizing Place Clusters and Reachability}~\label{supp:placegraph}
We provide a visualization of the place graphs from our test environment, derived purely from prior distribution and devoid of any updates.

\begin{figure*}[t!]{\centering\includegraphics[width=\textwidth]{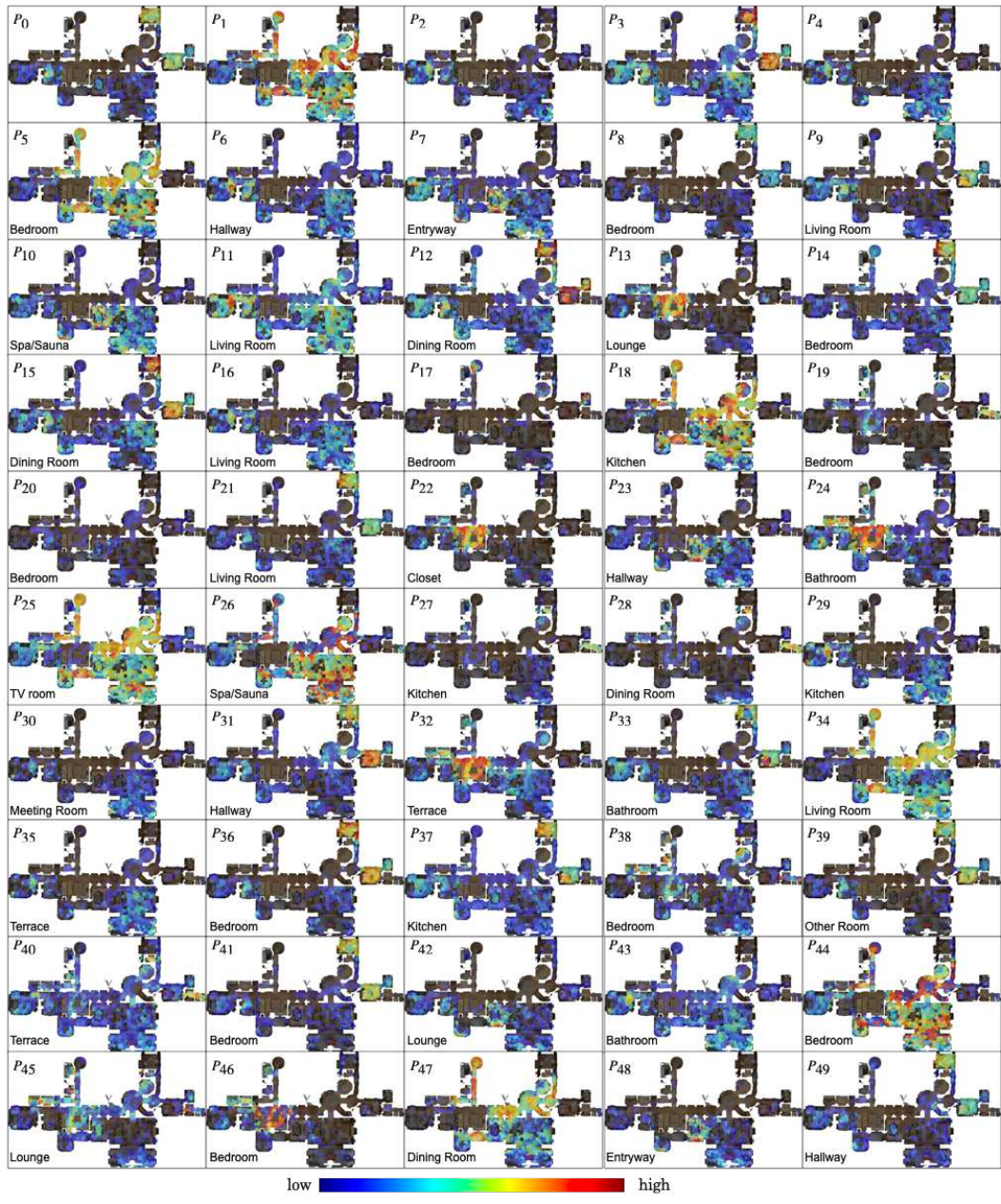}}\centering
\caption{{Visualization of place clusters on the map in a test environment (\textit{2azQ1b91cZZ}) of MP3D}}
\label{fig:cluster_2azQ1b91cZZ}
\end{figure*}

\begin{figure*}[t!]{\centering\includegraphics[width=\textwidth]{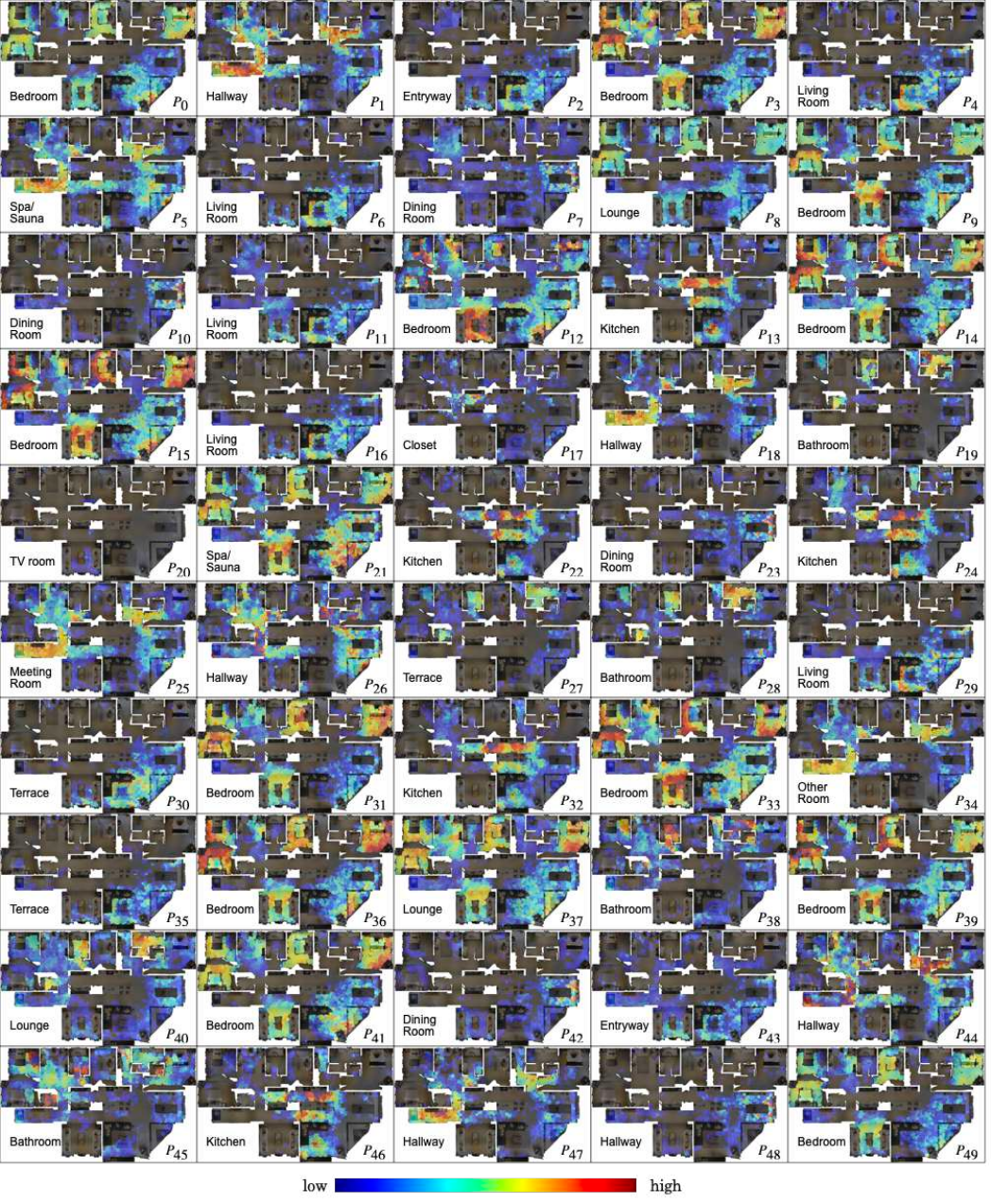}}\centering
\caption{{Visualization of place clusters on the map in a test environment (\textit{zsNo4HB9uLZ}) of MP3D}}
\label{fig:cluster_zsNo4HB9uLZ}
\end{figure*}

\paragraph{Place clusters}
Figure 5 of the manuscript illustrates an example of a semantic path using place clusters from MP3D. We further showcase examples of these place clusters in two test scenes from MP3D, depicted in Figures~\ref{fig:cluster_2azQ1b91cZZ} and ~\ref{fig:cluster_zsNo4HB9uLZ}.
These particular test scenes were chosen because they possess extensive areas, diverse rooms, and complex structures.
We employ RGB color-coding to represent the similarity between place clusters on the map, where red signifies high similarity and blue indicates low similarity.
Maps with a similarity score less than zero are not colored, thus displaying the original top-down RGB maps.
To compute the similarity between the place cluster centroid feature and the place features extracted from the map, features are gathered from 5000 random locations, as detailed in the experiment section of our main manuscript.
%
% The name of the place cluster is assigned based on the most closely matched place in the sample location, using the ground truth place.
%
Subsequently, the color coding serves to denote the likelihood of a place cluster's presence at a particular location on the map.
While the place names originate from the ground truth names acquired during training, the precise names of the place clusters remain undetermined.
It is crucial to mention that the place features and top-down maps are employed exclusively for analytical purposes and are not accessible to the agent.

\begin{figure*}[t!]{\centering\includegraphics[width=\textwidth]{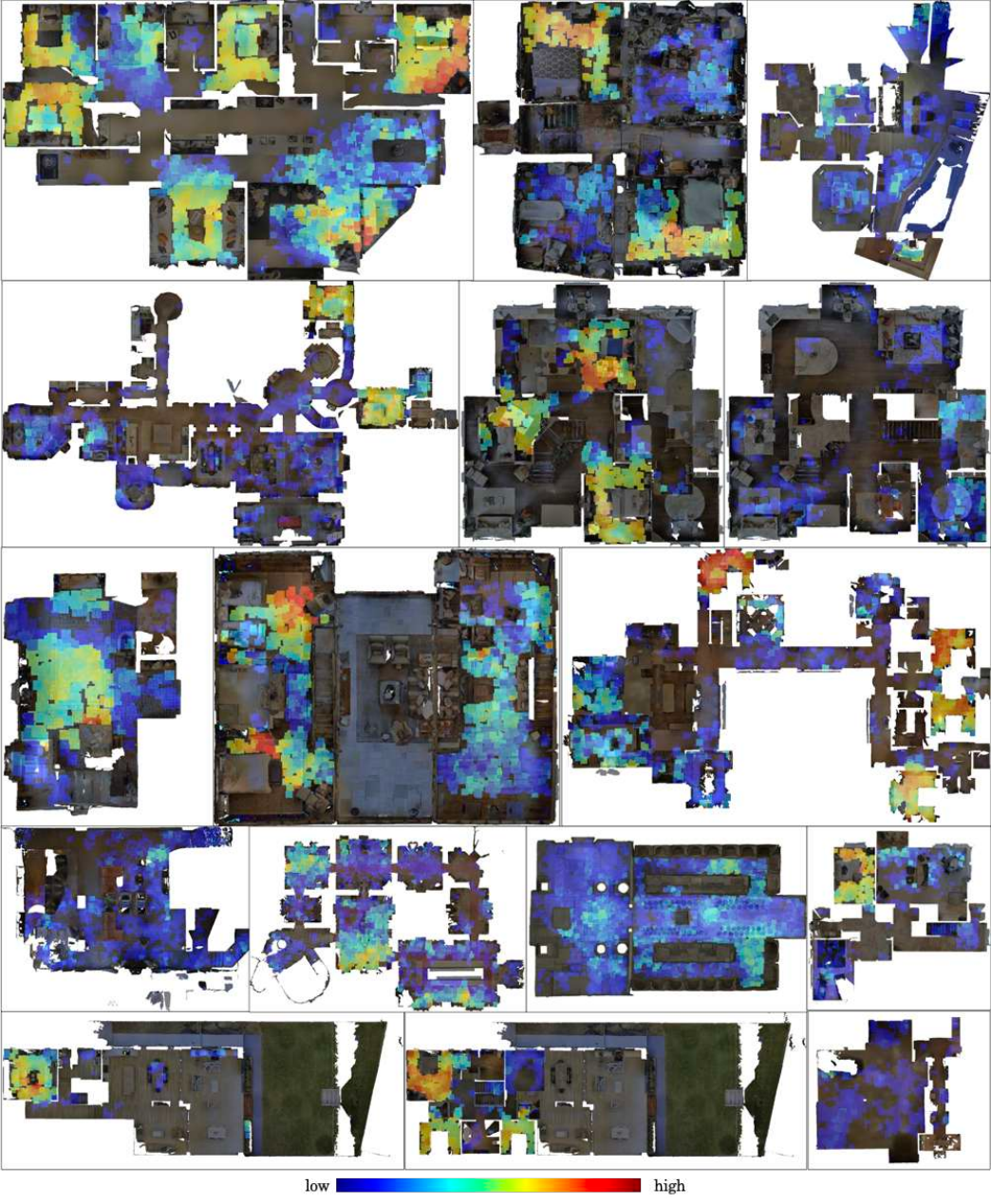}}\centering
\caption{{Visualization of a place cluster ($P_{41}$) on the maps in 11 test environments of MP3D}}
\label{fig:cluster_41}
\end{figure*}

% In Figure~\ref{fig:cluster_41}, we visualize the 41st place cluster (${P}_{41}$) across all 11 test scenes.
% Since some scene contains multiple floors, the number of maps is 16.
In Figure~\ref{fig:cluster_41}, we visualize the 41st place cluster (${P}_{41}$) across all 16 maps, which represent the 11 test scenes, taking into account that some scenes contain multiple floors.
% 
% The visualization shows that ${P}_{41}$ represents a bedroom, while the exact place names for place clusters are unknown since we used a clustering approach.
% The visualization shows that P41 represents a bedroom. However, the exact place names for other place clusters remain unknown due to our clustering approach.
We inferred the labels of the place clusters utilizing the ground truth place labels provided in the dataset. For instance, the 41st place cluster (${P}_{41}$) is representative of a bedroom.
% However, the exact place label for place clusters remain unknown because we use clustering approach.

\begin{figure*}[t!]{\centering\includegraphics[width=\textwidth]{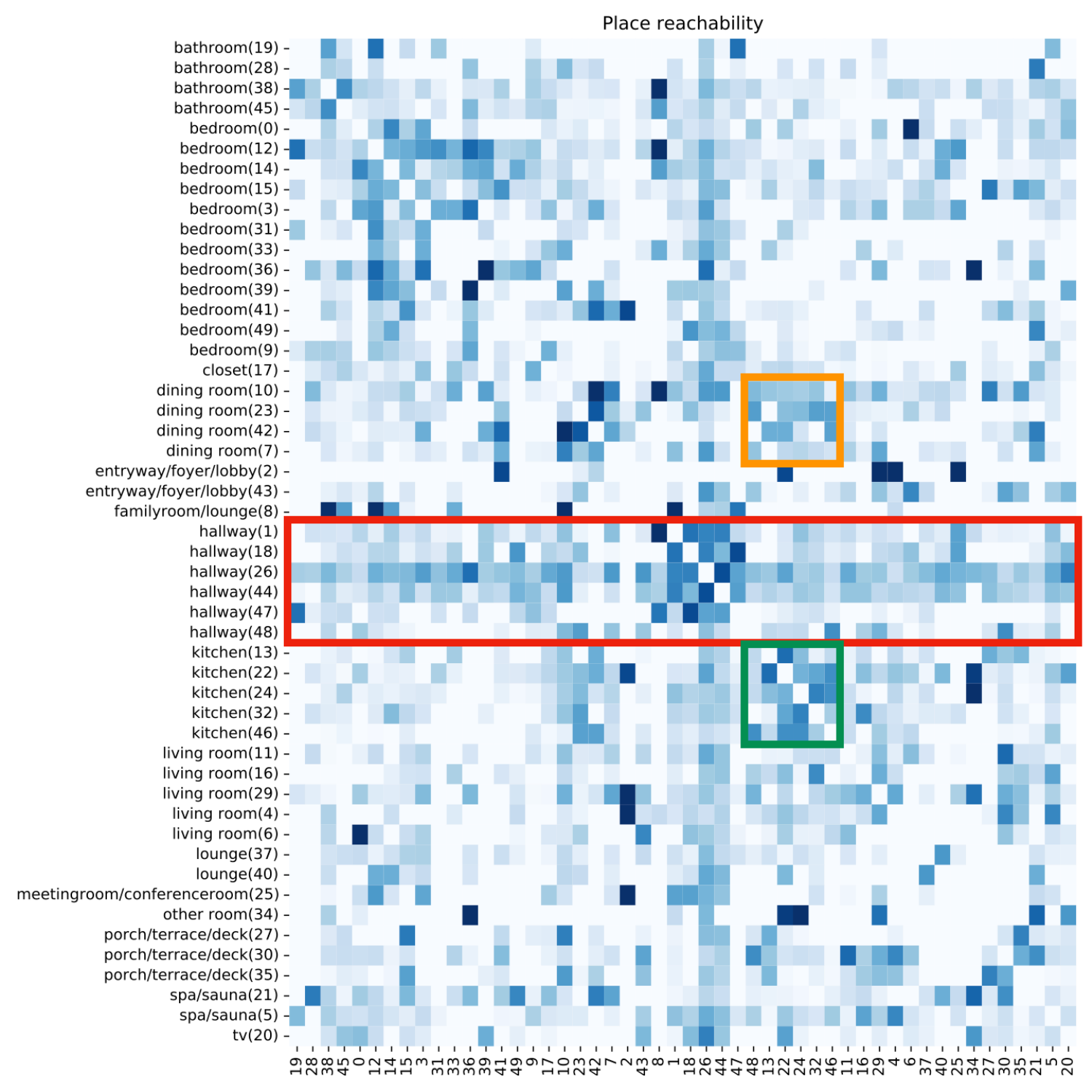}}\centering
\caption{ \textbf{Reachability of places.} The place clusters designated for the hallway, indicated by the {\color{darkred}\textbf{red}} box, have links to numerous other places, suggesting that the hallway serves as a bridge between different places.
The clusters associated with the kitchen are strongly connected to other kitchen types (denoted by the {\color{darkgreen}\textbf{green}} box) or dining rooms (highlighted in the {\color{darkorange}\textbf{orange}} box).
This illustrates that reachability is at one when place clusters are adjacent, whereas the connectivity between distant place clusters yields a reachability of zero.}
\label{fig:place_reachability_matrix}
\end{figure*}

\paragraph{Place reachability}
We show the reachability between place clusters in Figure~\ref{fig:place_reachability_matrix}, where the x-axis has been divided in half for visualization purposes only.
The reachability is zero if the two place clusters are not connected in the episodic graph. 
If two place clusters are observed in the environment and are always connected, the reachability is one.
As can be seen in Figure~\ref{fig:place_reachability_matrix}, the place clusters for the hallway (see {\color{darkred}\textbf{red}} box) are linked to many other places, which means that the hallway is a bridge between places.
The kitchen clusters are highly associated to other types of kitchens ({\color{darkgreen}\textbf{green}} box) or dining rooms ({\color{darkorange}\textbf{orange}} box).
It demonstrates that reachability is one when place clusters are adjacent and the connectivity between distant place clusters has zero reachability.
The meaningful reachability values have been collected using semantic graph maps.

\section{Visualization of ObjectNav Episodes}~\label{supp:Test}
\paragraph{Example visualization of episodes}
{\color{black}The navigation process in Figure~7 of the main manuscript} is as follow:
\begin{itemize}
    \item \textbf{Row 1.}  At first, the agent thinks the bedroom ($P_{12}$) is on the left side, and the kitchen is on the front and right side.
    The agent chooses to go to the left since the place on the left is expected to be a target place.
    \item \textbf{Row 2.} After reaching the subgoal, the agent looks for a bed, but it is not found.
    The agent exits the place and finds the subgoal again, and since the door is visible on the left, the agent can recognize that the target place will exist on the left.
    \item \textbf{Row 3.} Then, while going to the subgoal, the target object (bed) is found. The agent makes a local plan to reach the target object.
    At last, the agent presses the stop button since the place cluster matches the place searched for as the bedroom($P_{12}$).
\end{itemize}
Please check the supplemental video for more examples.

\paragraph{Trajectories on different goals}
{\color{black}Figure 8 in the main manuscript} illustrates trajectories for six object goals: fireplace, cabinet, chair, chest of drawers, bathtub, and bed. The trajectories corresponding to the six target object categories display a variety of pathways, each adapted to its respective target object:
\begin{itemize}
    \item \textbf{Fireplace.} The agent identifies the fireplace within the orange cluster. Initially, it searches the bedroom for the targeted object. Failing to find it in the bedroom, the agent proceeds to the next orange cluster, where it locates the fireplace.
    \item \textbf{Cabinet.} The agent recognizes that the cabinet is situated in the kitchen. It quickly locates the target by recognizing the kitchen in front of it.
    \item \textbf{Chair.} The agent discerns the presence of the chair in the orange cluster. Utilizing place reachability, it identifies the chair within the bedroom.
    \item \textbf{Chest of Drawers.} The agent encounters confusion as a chest of drawers is typically located in a bedroom (indicated by orange nodes), but in this environment, the living room contains the drawers. This discrepancy likely arises due to a mismatch between the test environment and the training environment. After some exploration, the agent successfully locates the target object, thanks to the adaptive property of the SEA.
    \item \textbf{Bathtub.} On determining that bathtubs are commonly located in bathrooms, the robot navigates its path to the bathroom, efficiently reaching the target area.
    \item \textbf{Bed.} The agent establishes that the bed is located in the bedroom. It approaches the corner of the scene via the hallway. Upon realizing there are no other rooms in the immediate vicinity, the agent proceeds to locate the bedroom elsewhere.
\end{itemize}

\vspace{11pt}

\clearpage
\clearpage
\clearpage
\clearpage
\clearpage

\clearpage

{
\bibliographystyle{elsarticle-num}
\bibliography{egbib}
}

\end{document}